\documentclass[11pt,letterpaper]{article}
\usepackage[utf8]{inputenc}
\usepackage[numbers]{natbib}
\usepackage{subcaption}
\DeclareUnicodeCharacter{202F}{\,}
\usepackage{makecell}
\usepackage{amsmath}
\usepackage{graphicx}
\usepackage{placeins}
\usepackage{epstopdf}
\usepackage{multirow}
\usepackage{times}
\usepackage{latexsym}
\usepackage{placeins}
\usepackage{amssymb}
\usepackage{newunicodechar}
\newunicodechar{，}{,}
\usepackage[T1]{fontenc}
\usepackage[utf8]{inputenc}
\usepackage{authblk}
\usepackage{bm}
\usepackage{booktabs}
\usepackage[table]{xcolor}  
\usepackage{colortbl}
\usepackage{microtype}
\usepackage{float}
\usepackage[lined,boxed,commentsnumbered]{algorithm2e}
\usepackage{inconsolata}
\usepackage[]{changes}
\usepackage{url}
\usepackage[T1]{fontenc}
\usepackage{newtxtext,newtxmath}
\usepackage{microtype}
\usepackage[margin=1in,top=0.88in,bottom=0.9in]{geometry}
\usepackage{xcolor}
\usepackage{graphicx}
\usepackage{array}
\usepackage[most]{tcolorbox}
\usepackage{fontawesome5}
\usepackage{titlesec}
\usepackage[colorlinks=true,linkcolor=tealbio,urlcolor=tealbio,citecolor=tealbio]{hyperref}
\usepackage{enumitem}

\usepackage[most]{tcolorbox}

\newcommand{\promptvar}[1]{%
    {\color{blue!75!black}\ttfamily\{#1\}}%
}

\newtcolorbox{biomemprompt}[1]{
    enhanced,
    colback=gray!5,
    colframe=black,
    boxrule=0.9pt,
    arc=9pt,
    left=12pt,
    right=12pt,
    top=17pt,
    bottom=11pt,
    fontupper=\small,
    title={#1},
    colbacktitle=black,
    coltitle=white,
    fonttitle=\bfseries\large,
    attach boxed title to top left={
        xshift=0pt,
        yshift=-\tcboxedtitleheight/2
    },
    boxed title style={
        colback=black,
        colframe=black,
        arc=9pt,
        left=12pt,
        right=12pt,
        top=4pt,
        bottom=4pt
    }
}

\newcommand{\TAMUARules}{%
\textbf{Task-Specific Evidence Utilization Rules}

\medskip
\textbf{Biomedical Question Answering}
\begin{itemize}[leftmargin=*,nosep]
    \item Identify biomedical entities and the relation
    requested by the question.
    \item Prioritize memories matching the same entities
    and relation, using their relevance ranking.
    \item Preserve the exact entity name, spelling,
    and word order when a matching answer exists.
    \item If no relevant memory matches, use biomedical
    knowledge and follow the expected entity naming style.
    \item Return only the requested entity names.
    Do not generate yes/no responses or explanations.
\end{itemize}

\medskip
\textbf{Clinical Risk Prediction}
\begin{itemize}[leftmargin=*,nosep]
    \item Compare the target patient with retrieved
    training patients using task-relevant diagnoses
    and clinical risk factors.
    \item Consider the training-set outcome prevalence
    \promptvar{prevalence} when interpreting class imbalance.
    \item Weight evidence by patient similarity rather than
    counting the majority outcome among retrieved memories.
    \item Return only \texttt{true} or \texttt{false}
    according to the task-specific prediction target.
\end{itemize}
}

\usepackage[most]{tcolorbox}
\usepackage{xcolor}
\usepackage{enumitem}

\newcommand{\rvar}[1]{%
    \textcolor{blue!75!black}{\{#1\}}%
}

\newtcolorbox{realmemPrompt}[1]{
    enhanced,
    colback=gray!4,
    colframe=black,
    boxrule=0.9pt,
    arc=9pt,
    left=12pt,
    right=12pt,
    top=17pt,
    bottom=11pt,
    fontupper=\footnotesize,
    title={#1},
    colbacktitle=black,
    coltitle=white,
    fonttitle=\bfseries\normalsize,
    attach boxed title to top left={
        xshift=0pt,
        yshift=-\tcboxedtitleheight/2
    },
    boxed title style={
        colback=black,
        colframe=black,
        arc=9pt,
        left=12pt,
        right=12pt
    }
}

\definecolor{tealbio}{HTML}{840028}
\definecolor{abstractfill}{HTML}{FCF7F8}
\definecolor{abstractedge}{HTML}{D9B8C0}
\definecolor{abstractheading}{HTML}{840028}

\setlist{nosep}

\titleformat{\section}{\large\bfseries}{\thesection}{0.7em}{}
\titleformat{\subsection}{\normalsize\bfseries}{\thesubsection}{0.7em}{}
\titlespacing*{\section}{0pt}{17pt}{7pt}
\titlespacing*{\subsection}{0pt}{12pt}{5pt}

\usepackage{graphicx}

\newcommand{\makebiomemtitle}{%
\begingroup
\setlength{\parskip}{0pt}
\setlength{\parindent}{0pt}

\begin{center}

{\LARGE\bfseries
BioMem: Agentic Memory for Biomedical LLM Agents\par}

\vspace{8pt}

\begin{center}

{\normalsize\bfseries
Mingchen Li\textsuperscript{1,4},
\quad
Yongkang Xiao\textsuperscript{3},
\quad
Rumeng Li\textsuperscript{2},
\quad
Rui Zhang\textsuperscript{3},
\quad
Hong Yu\textsuperscript{1,4,5}
\par}

\vspace{10pt}

{\footnotesize
\setlength{\parskip}{5pt}

\textsuperscript{1}\,
Manning College of Information and Computer Sciences,
University of Massachusetts Amherst
\par

\textsuperscript{2}\,
Department of Health Outcomes \& Biomedical Informatics,
University of Florida College of Medicine
\par

\textsuperscript{3}\,
Division of Computational Health Sciences,
Department of Surgery, University of Minnesota
\par

\textsuperscript{4}\,
VA Bedford Healthcare
\par

\textsuperscript{5}\,
Miner School of Computer \& Information Sciences,
University of Massachusetts Lowell
\par
}

\end{center}

\vspace{10pt}

\end{center}

\vspace{3pt}

\endgroup
}

\newtcolorbox{bioabstract}{
  colback=abstractfill,
  colframe=abstractedge,
  boxrule=0.65pt,
  arc=5pt,
  left=14pt,right=14pt,top=11pt,bottom=11pt,
  before skip=1pt,after skip=18pt
}

\begin{document}
\makebiomemtitle

\begin{bioabstract}
{\bfseries\color{abstractheading} Abstract}\par\vspace{3pt}
Large language models (LLMs) have demonstrated considerable potential in biomedical applications, while agentic memory systems have emerged as a promising approach to preserving and utilizing information across tasks and interactions. However, a fundamental question remains: how effectively and reliably do these systems retain, retrieve, update, and utilize biomedical information? In this study, we systematically investigate agentic memory in the biomedical domain, focusing on task effectiveness, retrieval quality, information fidelity, and long-term robustness.
We evaluate five memory approaches across four LLMs in three representative settings: biomedical knowledge reasoning, clinical prediction, and long-term conversational memory.
Our analyses reveal how memory construction, evidence relevance, and conversational history shape downstream biomedical performance, highlighting the distinct challenges of retrieving relevant information, preserving its fidelity, and effectively utilizing it for reasoning.
Building on these findings, we introduce Hybrid Memory Retrieval and Utilization (HMRU), a training-free framework that jointly enhances evidence accessibility and task-aware utilization to improve biomedical reasoning and long-term conversational memory.
HMRU achieves an average relative improvement of 8.6\% across  biomedical question-answering and conversational memory benchmarks, with additional gains in selected clinical prediction tasks. It further reduces conversational answer errors by 24.9\% and closes 48.0\% of the performance gap to full-context inference. Our findings provide a systematic characterization of biomedical agentic memory and practical insights into developing more effective and reliable memory-augmented LLM systems.
\end{bioabstract}

\section{Introduction}

Large language models (LLMs) have demonstrated considerable potential in biomedical applications, including clinical question answering, medical information extraction, and decision support \cite{li2025biomedrag,ma2026opbench,li2025benchmarking}. Advances in medical-domain pretraining and instruction tuning have further strengthened their ability to interpret biomedical terminology and respond to specialized clinical questions \cite{zhou2025accelerating,huang2026evolverouter,li2026cancerllm}. However, biomedical applications often depend on information that is patient-specific, evolving, or distributed across multiple sources, which cannot be fully captured by knowledge encoded in model parameters alone. These requirements motivate approaches that provide language models with access to external information beyond their parametric knowledge. Retrieval-augmented generation (RAG) addresses this need by incorporating retrieved evidence at inference time without updating the underlying model parameters~\cite{lewis2020retrieval}.

Beyond accessing external evidence, biomedical agents must also manage information accumulated through repeated interactions. Patient histories, evolving observations, and previous exchanges may remain relevant to subsequent tasks, requiring information to be preserved and updated over time. Persistent memory systems have emerged to address this need by organizing and reusing information across interactions~\cite{packer2023memgpt,chhikara2025mem0}. Recent approaches have further introduced specialized mechanisms for memory construction, retrieval, and updating.
A-MEM~\cite{xu2026mem}，for example, constructs interconnected memory notes and updates contextual representations as new information is incorporated . Mem0~\cite{chhikara2025mem0} extracts and consolidates salient conversational information for persistent, query-dependent retrieval, while LightMem~\cite{fang2026lightmem} separates lightweight online processing from longer-term consolidation to improve memory efficiency. These approaches illustrate different strategies for storing, structuring, and updating information that may be useful across tasks and interactions. Meanwhile, benchmarks such as RealMem~\cite{bian2026realmem} have begun to examine memory capabilities involving temporal reasoning, information updates, and interactions spanning multiple sessions.
However, the effectiveness of external memory systems depends on several interconnected processes. Information may be transformed during storage, selected or omitted during retrieval, revised through subsequent updates, and interpreted by the language model at inference time. This raises a fundamental question: \textit{how effectively and reliably do agentic memory systems preserve, retrieve, update, and utilize biomedical information?}

In this study, we systematically investigate agentic memory in biomedical domain, examining four dimensions: task effectiveness, retrieval and utilization quality, information fidelity, and long-term robustness. We evaluate five memory strategies with four language models across three representative settings: biomedical knowledge reasoning, clinical prediction, and long-term conversational memory. The evaluation covers six biomedical tasks, including reasoning over medical textual knowledge graphs , mortality and readmission prediction, cancer classification using, and opioid overdose prediction. 
We additionally evaluate long-term conversational memory tasks, examining incremental information, dynamic updating, proactive alignment, temporal reasoning, and cross-session interactions. Beyond overall task performance, we study the effects of memory construction data volume, retrieval depth, and class distribution. We also trace retrieved evidence to original records, assess its relevance to the target question, analyze source-grounded fidelity during iterative memory evolution, and evaluate robustness to conversation length and evidence distance.

Our analyses reveal four key findings. First, successful biomedical knowledge reasoning depends not only on retrieving relevant evidence but also on its appropriate utilization, as answer-bearing memories do not consistently lead to correct predictions. Second, semantic similarity between patient records does not necessarily correspond to similarity in future clinical outcomes, highlighting task-dependent limitations of memory retrieval. Third, iterative memory evolution is associated with the emergence and persistence of source-unsupported information, raising concerns about the fidelity of stored representations. Finally, long-term conversational memory performance depends on more than history length alone, with evidence accessibility and integration emerging as important factors. Together, these findings identify evidence retrieval, information fidelity, and memory utilization as distinct but interconnected challenges in biomedical agentic memory.
Motivated by these observations, we introduce Hybrid Memory Retrieval and Utilization (HMRU), a training-free framework designed to improve access to relevant evidence and its effective utilization in downstream biomedical tasks.
HMRU consistently improves performance on biomedical knowledge reasoning benchmarks and enhances long-term conversational memory by reducing answering errors and narrowing the performance gap with full-context inference. 
Our main contributions are summarized as follows:
\begin{enumerate}
    \item We systematically evaluate five memory methods using four LLMs across six biomedical tasks and long-term conversational memory settings, examining the effects of memory construction, retrieval depth, and class distribution on downstream performance.

    \item We investigate memory retrieval, evidence utilization, and information fidelity, identifying key limitations in biomedical reasoning and memory evolution. We further examine the effects of evidence completeness, conversation length, and evidence distance on long-term memory performance.

    \item We propose Hybrid Memory Retrieval and Utilization (HMRU), a training-free framework that enhances evidence retrieval and task-aware utilization, improving biomedical knowledge reasoning and long-term conversational memory performance.
\end{enumerate}

\section{Results}

\subsection{Memory Performance and Analysis on Biomedical Tasks}

Table~\ref{tab:training_data_memory} presents the performance of five memory methods across six biomedical benchmarks using four language models. The evaluation covers two biomedical knowledge reasoning tasks (RiTeK-PharmKG and RiTeK-ADint) and four clinical prediction and classification tasks (MIMIC-III Mortality, MIMIC-III Readmission, TCGA-Reports, and MIMIC-IV OOD). For knowledge reasoning tasks, we report precision, recall, and F1 scores using exact-match and ROUGE-1 measures. Clinical prediction tasks are evaluated using AUPRC, AUROC. We compare five memory management methods (A-MEM, MemZero, and LightMem, NaiveRAG and FullContext) to systematically examine how different memory mechanisms affect downstream biomedical task performance across model architectures and task settings. All results are reported as percentages.


\subsubsection{Comparisons with memory baselines}
\begin{table*}[!t]
\centering
\caption{Comparative evaluation of agentic memory approaches on biomedical benchmarks.
 Bold and underlined values indicate the best and second-best distinct results within each language model, respectively. For RiTeK datasets, $^*$ denotes exact match and $^\dagger$ denotes ROUGE-1.}
\label{tab:training_data_memory}

\scriptsize
\renewcommand{\arraystretch}{1.12}
\setlength{\tabcolsep}{2.6pt}

\resizebox{\textwidth}{!}{%
\begin{tabular}{@{}l*{20}{c}@{}}
\toprule
\multirow{2}{*}{\textbf{Method}}
& \multicolumn{6}{c}{\textbf{RiTeK-PharmKG}}
& \multicolumn{6}{c}{\textbf{RiTeK-ADInt}}
& \multicolumn{2}{c}{\textbf{Mort.-MIMIC-III}}
& \multicolumn{2}{c}{\textbf{Readm.-MIMIC-III}}
& \multicolumn{2}{c}{\textbf{TCGA-Reports}}
& \multicolumn{2}{c}{\textbf{MIMIC-IV OOD}} \\
\cmidrule(lr){2-7}
\cmidrule(lr){8-13}
\cmidrule(lr){14-15}
\cmidrule(lr){16-17}
\cmidrule(lr){18-19}
\cmidrule(lr){20-21}
& $P^*$ & $R^*$ & $F_1^*$
& $P^\dagger$ & $R^\dagger$ & $F_1^\dagger$
& $P^*$ & $R^*$ & $F_1^*$
& $P^\dagger$ & $R^\dagger$ & $F_1^\dagger$
& AUPRC & AUROC
& AUPRC & AUROC
& AUPRC & AUROC
& AUPRC & AUROC \\
\midrule

\rowcolor{gray!14}
\multicolumn{21}{@{}l}{\textit{Qwen3-8B}} \\

A-MEM
& \underline{26.97} & \underline{34.88} & \underline{28.53}
& \underline{29.25} & \textbf{37.76} & \underline{31.00}
& \underline{37.59} & \textbf{48.86} & \underline{40.17}
& \underline{41.94} & \textbf{55.81} & \underline{45.24}
& 9.61 & \underline{56.76}
& \textbf{59.25} & \underline{53.61}
& \underline{83.08} & \underline{97.88}
& 4.15 & \underline{84.21} \\

MemZero
& 16.56 & 26.28 & 18.92
& 18.70 & 29.00 & 21.19
& 23.81 & 38.76 & 27.59
& 27.42 & 45.41 & 31.98
& 8.98 & 53.07
& \underline{58.23} & 53.22
& 78.61 & 96.61
& 3.13 & 76.67 \\

FullContext
& 14.81 & 20.10 & 15.83
& 18.06 & 24.80 & 19.51
& 21.19 & 28.98 & 22.87
& 27.08 & 38.55 & 29.84
& 9.13 & 54.93
& 54.37 & 47.87
& 20.87 & 65.51
& 4.33 & 76.31 \\

NaiveRAG
& \textbf{28.34} & \textbf{35.08} & \textbf{29.57}
& \textbf{30.84} & \underline{37.75} & \textbf{32.16}
& \textbf{41.10} & \underline{48.20} & \textbf{42.74}
& \textbf{45.94} & \underline{54.56} & \textbf{48.15}
& \textbf{10.22} & \textbf{59.60}
& 57.79 & \textbf{53.75}
& 81.29 & 97.79
& \textbf{5.76} & 81.64 \\

LightMem
& 20.89 & 31.22 & 22.83
& 22.64 & 33.33 & 24.80
& 30.05 & 46.54 & 34.06
& 33.97 & 52.24 & 38.62
& \underline{10.14} & 56.16
& 55.40 & 50.57
& \textbf{88.64} & \textbf{98.67}
& \underline{4.84} & \textbf{84.71} \\

\midrule
\addlinespace[2pt]

\rowcolor{gray!14}
\multicolumn{21}{@{}l}{\textit{Meditron3-Qwen2.5-14B}} \\

A-MEM
& \underline{28.47} & \underline{32.13} & \underline{29.07}
& \underline{33.35} & \underline{37.55} & \underline{34.09}
& \underline{44.16} & \textbf{47.99} & \textbf{45.08}
& \underline{51.19} & \textbf{56.29} & \underline{52.59}
& 8.98 & 51.74
& \textbf{59.25} & \textbf{54.27}
& 77.58 & 97.19
& 2.49 & 69.53 \\

MemZero
& 21.36 & 24.00 & 21.87
& 27.93 & 31.17 & 28.56
& 30.69 & 33.82 & 31.51
& 38.83 & 43.86 & 40.18
& \textbf{13.53} & \textbf{60.22}
& 56.70 & 50.61
& \underline{86.15} & \underline{98.29}
& 2.20 & \underline{72.13} \\

FullContext
& 19.77 & 21.77 & 19.90
& 23.32 & 26.24 & 23.75
& 28.82 & 31.19 & 29.22
& 37.77 & 41.61 & 38.61
& 7.22 & 48.60
& 54.65 & 49.25
& 24.13 & 73.13
& \underline{2.86} & 70.60 \\

NaiveRAG
& \textbf{30.69} & \textbf{34.15} & \textbf{31.23}
& \textbf{35.47} & \textbf{39.15} & \textbf{36.01}
& \textbf{44.23} & \underline{47.48} & \underline{44.90}
& \textbf{51.91} & \underline{55.82} & \textbf{52.89}
& 8.55 & 55.75
& \underline{57.55} & \underline{53.65}
& 76.25 & 97.39
& \textbf{3.58} & \textbf{73.69} \\

LightMem
& 26.52 & 28.92 & 26.93
& 31.50 & 34.17 & 32.00
& 38.12 & 41.67 & 38.94
& 45.06 & 50.01 & 46.42
& \underline{10.51} & \underline{56.18}
& 54.98 & 48.42
& \textbf{89.05} & \textbf{98.68}
& 2.64 & 68.07 \\

\midrule
\addlinespace[2pt]

\rowcolor{gray!14}
\multicolumn{21}{@{}l}{\textit{HuatuoGPT-3-27B}} \\

A-MEM
& \textbf{39.95} & \underline{39.23} & \textbf{39.27}
& \textbf{42.18} & \textbf{41.40} & \textbf{41.51}
& \textbf{53.32} & \textbf{53.08} & \textbf{52.94}
& \textbf{58.84} & \textbf{58.62} & \textbf{58.53}
& \textbf{9.81} & \underline{59.88}
& \textbf{61.91} & \textbf{56.69}
& \textbf{99.06} & \textbf{99.93}
& 4.94 & 82.16 \\

MemZero
& 31.22 & 31.07 & 30.84
& 33.96 & 33.76 & 33.57
& 43.57 & 43.43 & 43.31
& 50.05 & 49.99 & 49.84
& 8.29 & 54.21
& 57.98 & 52.71
& 98.64 & 99.87
& 6.95 & 78.94 \\

FullContext
& 27.55 & 26.65 & 26.80
& 30.25 & 29.51 & 29.64
& 37.34 & 36.58 & 36.83
& 45.30 & 44.68 & 44.90
& \underline{9.68} & 59.19
& 57.93 & 52.55
& 98.88 & \underline{99.88}
& \underline{8.23} & \underline{82.76} \\

NaiveRAG
& \underline{39.42} & \textbf{39.33} & \underline{39.04}
& \underline{41.48} & \underline{41.36} & \underline{41.07}
& \underline{52.07} & \underline{51.63} & \underline{51.63}
& \underline{57.69} & \underline{57.34} & \underline{57.36}
& \textbf{9.81} & \textbf{61.98}
& \underline{60.76} & \underline{55.79}
& \underline{98.97} & \textbf{99.93}
& \textbf{15.61} & \textbf{85.47} \\

LightMem
& 37.05 & 36.84 & 36.62
& 39.25 & 38.89 & 38.78
& 50.21 & 49.76 & 49.76
& 55.06 & 54.88 & 54.77
& 9.62 & 57.80
& 59.17 & 53.91
& 98.47 & 99.81
& 3.42 & 79.11 \\

\midrule
\addlinespace[2pt]

\rowcolor{gray!14}
\multicolumn{21}{@{}l}{\textit{Qwen3-30B-A3B-Instruct-2507}} \\

A-MEM
& \underline{31.79} & \underline{36.23} & \underline{33.00}
& \underline{33.95} & \underline{38.64} & \underline{35.28}
& \underline{44.35} & \textbf{50.80} & \underline{46.20}
& \underline{50.69} & \textbf{58.69} & \underline{53.18}
& \underline{9.37} & 51.17
& 55.01 & 49.76
& \textbf{98.39} & \underline{99.81}
& \textbf{4.13} & \textbf{81.66} \\

MemZero
& 23.76 & 29.20 & 25.29
& 25.26 & 31.31 & 27.01
& 28.75 & 35.79 & 30.80
& 34.82 & 44.95 & 37.86
& 7.54 & 51.35
& \underline{57.13} & \underline{52.33}
& 96.76 & 99.75
& 2.67 & 74.60 \\

FullContext
& 24.11 & 26.16 & 24.42
& 26.53 & 29.12 & 27.07
& 29.96 & 34.09 & 31.04
& 37.46 & 44.43 & 39.57
& \textbf{10.27} & \underline{52.90}
& 55.50 & 49.78
& 95.41 & 99.43
& 3.52 & 73.96 \\

LightMem
& 29.67 & 33.77 & 30.73
& 31.31 & 35.57 & 32.47
& 39.64 & 45.30 & 41.16
& 45.29 & 52.65 & 47.49
& 8.88 & 52.19
& 54.76 & 49.76
& 97.76 & \underline{99.81}
& 3.25 & 77.61 \\

NaiveRAG
& \textbf{35.24} & \textbf{40.29} & \textbf{36.53}
& \textbf{36.87} & \textbf{41.87} & \textbf{38.18}
& \textbf{46.14} & \underline{50.17} & \textbf{47.29}
& \textbf{52.68} & \underline{57.93} & \textbf{54.26}
& 9.02 & \textbf{56.57}
& \textbf{57.89} & \textbf{54.70}
& \underline{98.36} & \textbf{99.90}
& \underline{3.92} & \underline{78.59} \\

\bottomrule
\end{tabular}%
}
\end{table*}

We evaluated five memory approaches across six biomedical benchmarks using four language models (Table~\ref{tab:training_data_memory}). Performance varied substantially across tasks, memory strategies, and model backbones. In biomedical knowledge reasoning, A-MEM and NaiveRAG generally achieved stronger performance than MemZero, LightMem, and FullContext. With HuatuoGPT-3-27B, A-MEM achieved exact-match F1 scores of 39.27 and 52.94 on RiTeK-PharmKG and RiTeK-ADInt, respectively, compared with 39.04 and 51.63 for NaiveRAG and 26.80 and 36.83 for FullContext. However, this advantage was not consistent across model backbones. With Qwen3-30B, NaiveRAG outperformed A-MEM on both datasets, achieving F1 scores of 36.53 versus 33.00 on PharmKG and 47.29 versus 46.20 on ADInt. 
Clinical prediction and pathology classification exhibited different performance patterns. On TCGA-Reports, LightMem achieved the highest AUPRC of 88.64 with Qwen3-8B and 89.05 with Meditron3-Qwen2.5-14B, whereas A-MEM achieved an AUPRC of 99.06 and an AUROC of 99.93 with HuatuoGPT-3-27B. For MIMIC-III Mortality, MemZero achieved the highest AUPRC of 13.53 with Meditron3-Qwen2.5-14B, outperforming A-MEM and NaiveRAG, which achieved AUPRC values of 8.98 and 8.55, respectively. In contrast, NaiveRAG showed a pronounced advantage on MIMIC-IV opioid overdose prediction with HuatuoGPT-3-27B, achieving an AUPRC of 15.61 and an AUROC of 85.47, compared with an AUPRC of 4.94 and an AUROC of 82.16 for A-MEM. On MIMIC-III Readmission, AUPRC values were relatively high despite AUROC values remaining close to chance level across several configurations, indicating limited discrimination and emphasizing the importance of interpreting precision--recall performance in relation to outcome prevalence.

\FloatBarrier

\subsubsection{Factors Influencing Memory Performance}

\begin{figure*}[!t]
    \centering
    \includegraphics[width=\textwidth]{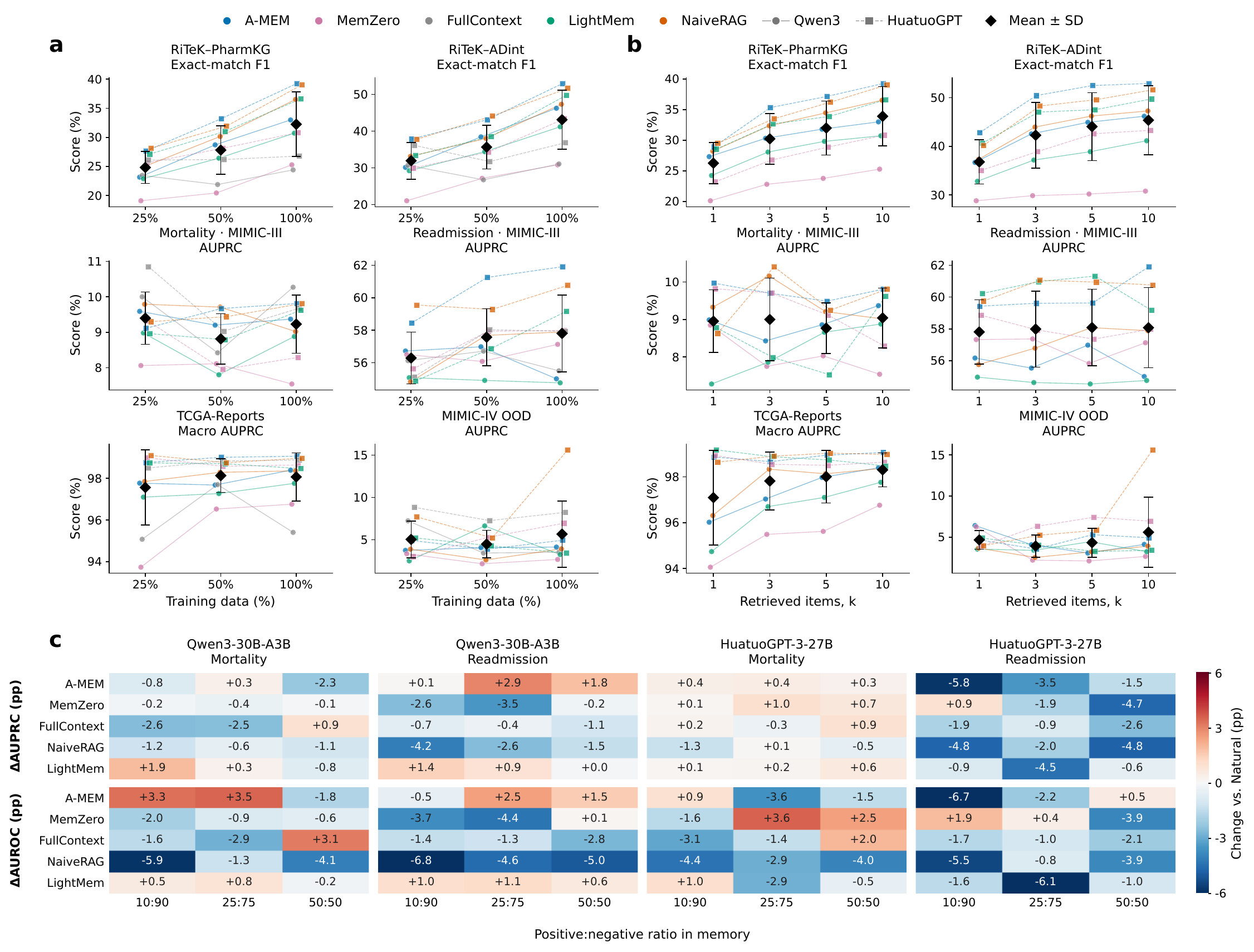}
    \caption{\textbf{Combined Sensitivity (test).}
    \textbf{a}, Effect of training data volume.
    \textbf{b}, Effect of retrieval depth (Top-k) on biomedical memory performance.
    \textbf{c}, Effect of class distribution on biomedical memory performance.}
    \label{fig:combined_sensitivity}
\end{figure*}

To better understand the factors influencing memory performance in biomedical tasks, we systematically investigate three key aspects: training data volume, retrieval depth, and class distribution. First, we examine how varying the number of training samples used for memory construction affects downstream performance. Second, we evaluate the impact of retrieval depth by varying the number of memory items retrieved during inference. Third, we investigate how different positive-to-negative class distributions in memory construction influence clinical prediction performance. Across these experiments, we aim to characterize the sensitivity of different memory methods to data availability, retrieval configurations, and class imbalance, providing insights into their effectiveness and robustness across biomedical tasks.

\paragraph{Effect of training data volume.}

Figure~\ref{fig:combined_sensitivity}(a) illustrates the effects of memory construction data volume across six biomedical tasks, using 250, 500, and 1,000 samples. Increasing data volume generally improves performance on biomedical knowledge reasoning tasks, whereas clinical prediction and pathology classification exhibit more heterogeneous patterns. On RiTeK-PharmKG, A-MEM with HuatuoGPT-3-27B achieves exact-match F1 scores of 27.66, 33.23, and 39.27 as data volume increases, while corresponding scores on RiTeK-ADint rise from 37.83 to 43.08 and 52.94. Similar improvements are observed with Qwen3-30B, where A-MEM increases from 23.17 to 33.00 on RiTeK-PharmKG. In contrast, performance on clinical prediction tasks does not consistently improve with additional data. On MIMIC-IV OOD, NaiveRAG with HuatuoGPT-3-27B achieves AUPRC scores of 7.70, 5.22, and 15.61, whereas LightMem decreases from 5.21 to 4.28 and 3.42. Performance on TCGA-Reports remains comparatively stable across data volumes, particularly for configurations with already high AUPRC values. 



\paragraph{Effect of retrieval depth.}
Figure~\ref{fig:combined_sensitivity}(b) illustrates the effects of retrieval depth across six biomedical tasks, with the number of retrieved memory items varying from 1 to 10. Increasing retrieval depth generally improves performance on biomedical knowledge reasoning tasks, whereas clinical prediction and pathology classification exhibit more heterogeneous patterns. On RiTeK-PharmKG, A-MEM with Qwen3-30B achieves exact-match F1 scores of 27.33, 30.32, 31.81, and 33.00 at Top-1, Top-3, Top-5, and Top-10, respectively. Similar improvements are observed on RiTeK-ADint, where exact-match F1 increases from 36.71 to 46.20. In contrast, deeper retrieval does not consistently improve clinical prediction. On MIMIC-IV OOD, NaiveRAG with HuatuoGPT-3-27B achieves an AUPRC of 3.99 at Top-1 and 15.61 at Top-10, while other configurations show smaller gains or fluctuating performance. On MIMIC-III Readmission, LightMem reaches its highest AUPRC of 61.32 at Top-5 before declining to 59.17 at Top-10. Performance on TCGA-Reports remains relatively stable at higher retrieval depths, particularly for configurations with already high AUPRC values. 


\paragraph{Effect of class distribution.}
Figure~\ref{fig:combined_sensitivity} (c) illustrates the effects of varying positive-to-negative class ratios in memory across two clinical prediction tasks. Changes in class distribution produce heterogeneous effects on predictive performance, with no single ratio consistently outperforming the natural distribution across memory methods and language models. On MIMIC-III Readmission, A-MEM with Qwen3-30B achieves an AUPRC of 57.92 under the 25:75 ratio, compared with 55.01 under the natural distribution. Similarly, LightMem with Qwen3-30B improves mortality AUPRC from 8.88 to 10.77 under the 10:90 ratio. However, increasing the proportion of positive samples does not consistently improve performance. With HuatuoGPT-3-27B, A-MEM achieves its highest readmission AUPRC of 61.91 under the natural distribution, declining to 60.37 under the 50:50 ratio. NaiveRAG also exhibits substantial performance reductions across several configurations, particularly for readmission prediction.

\subsubsection{Memory Retrieval Quality and Utilization}

\begin{figure*}[!t]
    \centering
    \includegraphics[width=\textwidth]{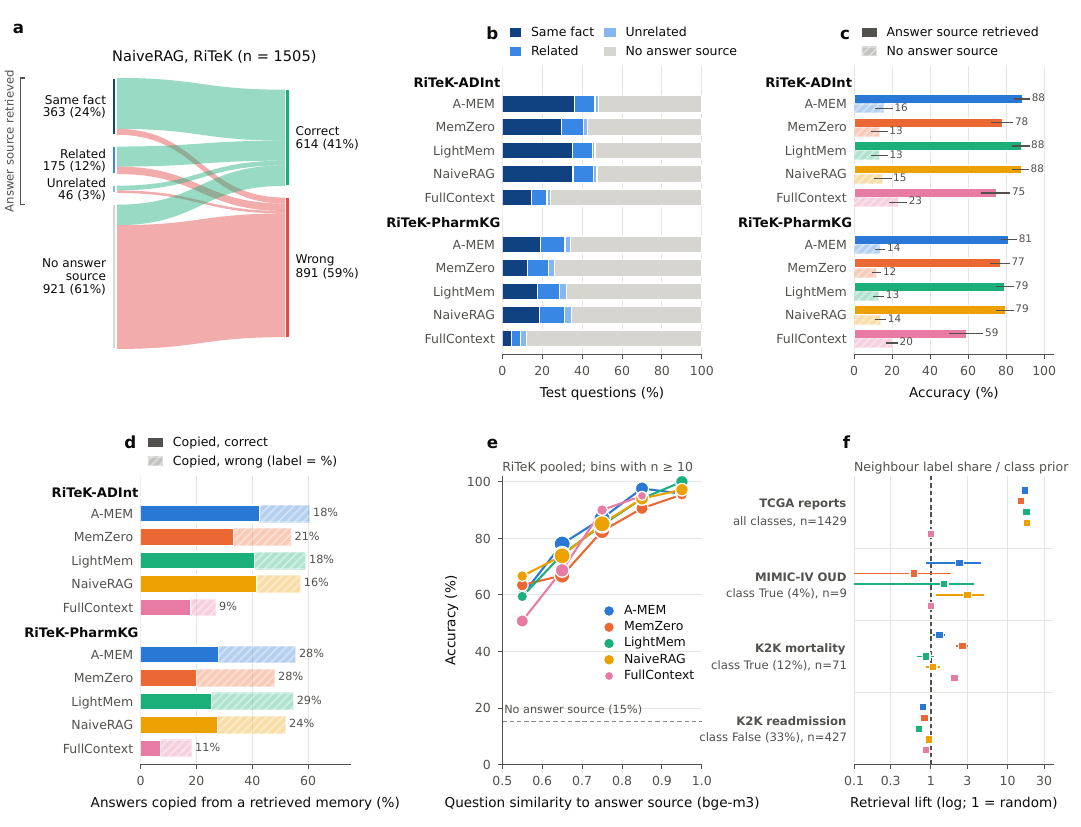}
    \caption{\textbf{Memory retrieval quality and utilization across biomedical tasks.}
    \textbf{a}, Evidence-to-outcome flows for NaiveRAG on the pooled RiTeK test questions.
    \textbf{b}, Distribution of the strongest retrieved evidence category across five memory methods.
 \textbf{c}， Answer accuracy stratified by whether retrieved memory contains an answer matching the ground truth.
    \textbf{d}, Proportions of predictions matching a retrieved source answer, separated by correct and incorrect predictions.
    \textbf{e}, Answer accuracy by BGE-M3 similarity between the test question and the strongest answer-matching source question; bins contain at least ten questions.
    \textbf{f}, Retrieval enrichment for the ground-truth class relative to its training-set prior in four classification tasks; error bars show 95 per cent confidence intervals.}
    \label{fig:memory_layers}
\end{figure*}


To investigate how memory retrieval and utilization influence downstream biomedical performance, we systematically analyzed retrieved evidence and model predictions across six biomedical benchmarks (Fig.~\ref{fig:memory_layers}a--f). For biomedical question answering, we examined the factual relevance of retrieved sources, the association between answer-source availability and prediction accuracy, the extent to which model outputs matched retrieved source answers, and the relationship between evidence similarity and answering performance. For clinical prediction and classification, we assessed whether retrieved examples were enriched for the ground-truth class relative to its prevalence in the training data. 

\paragraph{Relationship between retrieved evidence and answer correctness.}

Figure~\ref{fig:memory_layers}a illustrates the relationship between retrieved evidence categories and answer correctness across the two RiTeK benchmarks using NaiveRAG. Retrieved evidence is categorized as containing the same fact, related information, unrelated information with a matching answer, or no answer-matching source. Among the 1,505 questions, 363 contain a same-fact source, 175 a related source, and 46 an unrelated source with a matching answer, accounting for 24\%, 12\%, and 3\% of the sample, respectively. The remaining 921 questions, comprising 61\% of the sample, lack an answer-matching source. Overall, NaiveRAG correctly answers 614 questions and incorrectly answers 891, yielding an accuracy of 41\%. Correct answers are more strongly associated with factually relevant evidence, although errors persist even when answer-bearing sources are retrieved. Conversely, some questions are answered correctly without an identifiable answer-matching source. These findings highlight the relationship between evidence availability, factual relevance, and answering performance, while indicating that retrieving relevant evidence alone does not guarantee correct utilization.

\paragraph{Coverage of answer-bearing and factually relevant sources.}
Figure~\ref{fig:memory_layers}\textbf{b} compares the distribution of retrieved evidence categories across memory methods and RiTeK datasets. For each method and dataset, the proportion assigned to evidence category $c$ is defined as $P(c)=\frac{1}{N}\sum_{i=1}^{N}\mathbb{1}[C_i=c]$, where $c\in\{\mathrm{same},\mathrm{related},\mathrm{unrelated},\mathrm{none}\}$，
where $C_i$ is the highest evidence category assigned to test question $i$. On RiTeK-ADInt, A-MEM, LightMem, and NaiveRAG retrieve a same-fact source for approximately 35 per cent of questions, compared with approximately 30 per cent for MemZero and 15 per cent for FullContext. On RiTeK-PharmKG, the corresponding values decrease to approximately 18 per cent for the first three systems, 13 per cent for MemZero, and 5 per cent for FullContext. The larger no-answer-source segments on PharmKG indicate weaker answer-source coverage. Importantly, more elaborate memory construction does not exhibit a clear coverage advantage over NaiveRAG in these traces. 

\paragraph{Accuracy conditional on answer-source retrieval.}
Figure~\ref{fig:memory_layers}\textbf{c} compares answer accuracy across memory methods according to whether an answer-matching source is retrieved. For $a\in\{0,1\}$, conditional accuracy is defined as
\begin{equation}
\operatorname{Acc}(A=a)=
\frac{\sum_{i=1}^{N}\mathbb{1}[A_i=a]\mathbb{1}[\hat y_i=y_i]}
{\sum_{i=1}^{N}\mathbb{1}[A_i=a]},
\label{eq:memory_conditional_accuracy}
\end{equation}
where $A_i$ indicates whether an answer-matching source is retrieved for question $i$. Solid and hatched bars represent accuracy with and without an answer-matching source, respectively. Across the four retrieval-based methods, accuracy ranges from 77\% to 88\% when an answer-matching source is available, compared with only 12\% to 16\% when no such source is retrieved. On RiTeK-ADInt, A-MEM achieves an accuracy of 88\% with an answer-matching source, compared with 16\% without one. Similarly, NaiveRAG achieves 88\% and 15\%, respectively. FullContext exhibits a smaller separation, with conditional accuracies of 75\% and 23\% on RiTeK-ADInt, and 59\% and 20\% on RiTeK-PharmKG. Although accuracy is substantially higher when answer-matching evidence is available, incorrect predictions remain even when such evidence is retrieved. These findings demonstrate a strong association between answer-source availability and prediction accuracy, while indicating that successful retrieval does not guarantee correct evidence utilization.

\paragraph{Correct and incorrect matching of retrieved source answers.}
Figure~\ref{fig:memory_layers}d examines the extent to which model predictions match answers contained in retrieved memory sources. For each test question, we define $B_i=\mathbb{1}[\operatorname{norm}(\hat y_i)\in\{\operatorname{norm}(y_s):s\in\mathcal S(m),\,m\in\mathcal M_i\}]$, where $B_i=1$ indicates that the predicted answer matches at least one retrieved source answer. We further distinguish correct and incorrect predictions using the proportion $P(\mathrm{matched\;and\;wrong})=\frac{1}{N}\sum_{i=1}^{N}B_i\,\mathbb{1}[\hat y_i\neq y_i]$, calculated over all test questions.
Across the four retrieval-based methods, incorrect source-answer matches account for approximately one-fifth of questions on RiTeK-ADInt and one-quarter on RiTeK-PharmKG. In contrast, the corresponding proportions for FullContext are 9\% and 11\%, respectively. These findings indicate that incorrect predictions frequently coincide with answers present in retrieved memory, suggesting potential interference from irrelevant or mismatched evidence.

\paragraph{Evidence similarity and answer accuracy.}
Figure~\ref{fig:memory_layers}\textbf{e} examines the association between source-question similarity and answer accuracy. For questions with at least one answer-matching source ($A_i=1$), we calculated the maximum cosine similarity $s_i^*=\max_{s\in\mathcal S_i^+}\operatorname{sim}(\mathbf e(q_i),\mathbf e(q_s))$, where $\mathcal S_i^+$ denotes the retrieved source records whose answers match the reference answer, $\mathbf e$ is the BGE-M3 text encoder, and $\operatorname{sim}$ denotes cosine similarity. Questions from both RiTeK datasets were pooled and stratified by similarity, excluding bins containing fewer than ten observations. Answer accuracy increased progressively with source-question similarity, approaching 100 per cent among questions with the highest similarity scores. This relationship was consistent across all five memory methods, with largely overlapping accuracy trajectories. In contrast, questions without an answer-matching source exhibited substantially lower accuracy.

\paragraph{Retrieval enrichment for classification outcomes.}
Figure~\ref{fig:memory_layers} \textbf{f} examines whether retrieved training examples are enriched for the ground-truth class of each test instance relative to its prevalence in the training data. For each test instance $i$, we calculate the fraction of retrieved examples sharing its label as $f_i=\frac{1}{k_i}\sum_{j=1}^{k_i}\mathbb{1}[y_{ij}=y_i]$, where $k_i$ denotes the number of retrieved source labels. For a group of test instances $G$, retrieval lift is defined as $\operatorname{Lift}(G)=\frac{\sum_{i\in G}f_i}{\sum_{i\in G}\pi_{y_i}}$, where $\pi_{y_i}$ is the training-set prevalence of the ground-truth class. A lift of one corresponds to the expected same-label retrieval rate under random sampling from the training distribution.
On TCGA-Reports, the four retrieval-based methods exhibited substantial enrichment of same-class pathology reports, with retrieval lift exceeding 15, whereas FullContext remained close to the class-prior baseline. In contrast, enrichment was considerably weaker and more heterogeneous across clinical outcome prediction tasks. On MIMIC-III Mortality, MemZero achieved a lift of 2.6 for positive cases, while the remaining methods showed limited enrichment. For MIMIC-III Readmission, retrieval lift for the minority negative class was generally below one, indicating that retrieved examples did not consistently reflect the test instances' outcome labels. Estimates for MIMIC-IV opioid overdose prediction were highly uncertain because only nine positive test instances were available.



\FloatBarrier

\subsubsection{Memory evolution and fidelity.}

\begin{figure*}[!t]
    \centering
    \includegraphics[width=\textwidth]{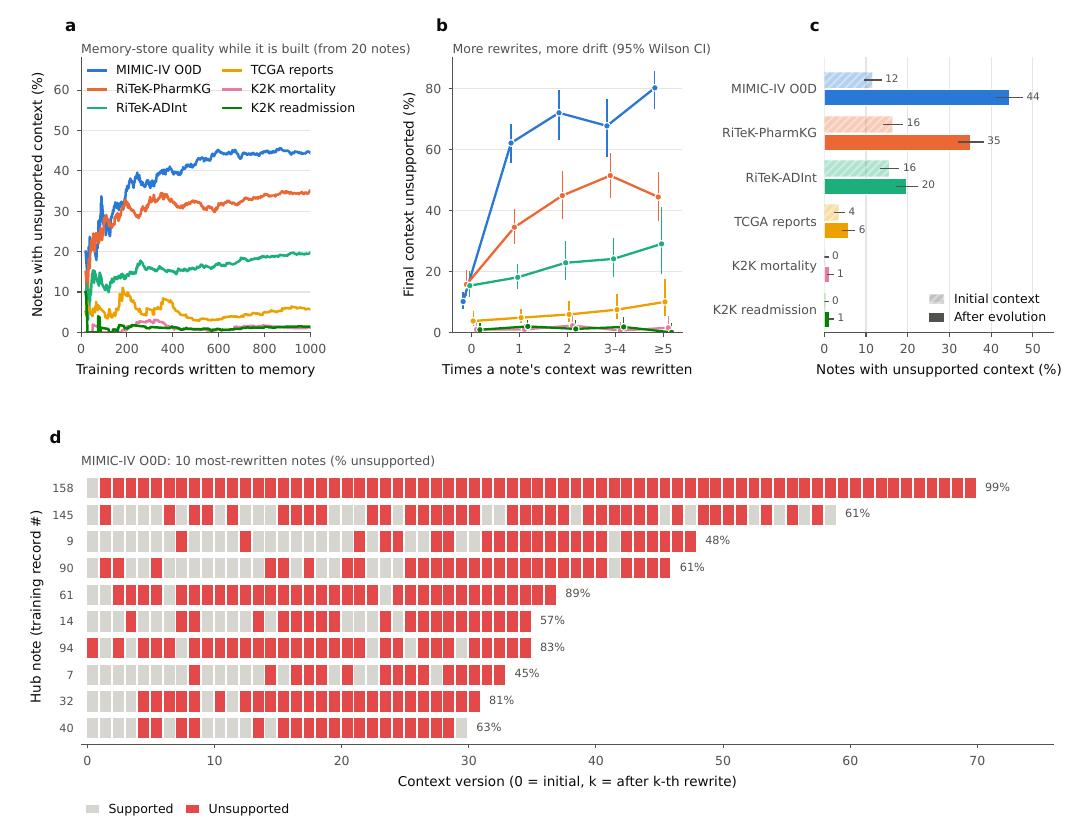}
    \caption{\textbf{Memory evolution in A-MEM.}
    \textbf{a}, Proportion of memory nodes with unsupported context during sequential memory construction across six biomedical tasks.
    \textbf{b}, Association between context rewrite frequency and unsupported context rate, with 95 per cent Wilson confidence intervals.
    \textbf{c}, Proportion of memory notes containing source-unsupported context in their initial and final states after memory evolution.
    \textbf{d}, Evolution trajectories of the ten most frequently rewritten nodes in MIMIC-IV OOD. Gray and red indicate supported and unsupported context, respectively. Unsupported content is determined by an LLM evaluator against the original training record.}
    \label{fig:amem_evolution}
\end{figure*}

We further investigated information fidelity during A-MEM construction and evolution across six biomedical tasks (Fig.~\ref{fig:amem_evolution}). Using an LLM-based evaluator, we assessed whether generated memory contexts contained information unsupported by their original training records. During sequential memory construction, the proportion of memories containing unsupported context increased in several tasks, reaching 44\% in MIMIC-IV OOD and 35\% in RiTeK-PharmKG, while remaining below 2\% in both MIMIC-III tasks (Fig.~\ref{fig:amem_evolution}a). Higher rewrite frequency was associated with a greater prevalence of unsupported information, particularly in MIMIC-IV OOD, where the proportion increased from approximately 10\% among memories without rewrites to 80\% among those rewritten at least five times (Fig.~\ref{fig:amem_evolution}b).
Comparisons between initial memory contexts and their final states after evolution revealed substantial task-dependent differences (Fig.~\ref{fig:amem_evolution}c). In MIMIC-IV OOD, the proportion of memories containing unsupported context increased from 12\% to 44\%, representing a 32-percentage-point increase. RiTeK-PharmKG showed a similar pattern, increasing from 16\% to 35\%. In contrast, smaller changes were observed in RiTeK-ADInt and TCGA-Reports, while both MIMIC-III tasks remained close to zero throughout memory evolution. Longitudinal inspection of the ten most frequently rewritten memories in MIMIC-IV OOD further revealed extended sequences of unsupported context, although some memories alternated between supported and unsupported states across successive revisions (Fig.~\ref{fig:amem_evolution}d).

\FloatBarrier

\subsection{Biomedical Conversational Memory}

\subsubsection{Overall Performance of Biomedical Conversational Memory}

\begin{table*}[!t]
\centering
\caption{\textbf{Memory performance on the RealMem medical subset.}
Performance of five memory methods across four language-model
backbones, four memory categories, and two session types.
QA and Rec. denote QA Score and Memory Recall, respectively.
Scores are percentages. Bold and underlined values indicate
the best and second-best distinct scores within each backbone
and metric. Dyn. Inc., dynamic incremental; Dyn. Upd.,
dynamic updating; Proac. Align., proactive alignment;
Temp. Reas., temporal reasoning.}
\label{tab:medical_memory_category}

\scriptsize
\renewcommand{\arraystretch}{1.12}
\setlength{\tabcolsep}{2.8pt}

\resizebox{0.75\textwidth}{!}{%
\begin{tabular}{@{}l*{12}{c}@{}}
\toprule
\multirow{3}{*}{\textbf{Method}}
& \multicolumn{8}{c}{\textbf{Memory Category}}
& \multicolumn{4}{c}{\textbf{Session Type}} \\
\cmidrule(lr){2-9}
\cmidrule(lr){10-13}

& \multicolumn{2}{c}{\textbf{Dyn. Inc.}}
& \multicolumn{2}{c}{\textbf{Dyn. Upd.}}
& \multicolumn{2}{c}{\textbf{Proac. Align.}}
& \multicolumn{2}{c}{\textbf{Temp. Reas.}}
& \multicolumn{2}{c}{\textbf{Single Sess.}}
& \multicolumn{2}{c}{\textbf{Multi Sess.}} \\

\cmidrule(lr){2-3}
\cmidrule(lr){4-5}
\cmidrule(lr){6-7}
\cmidrule(lr){8-9}
\cmidrule(lr){10-11}
\cmidrule(lr){12-13}

& QA & Rec. & QA & Rec.
& QA & Rec. & QA & Rec.
& QA & Rec. & QA & Rec. \\
\midrule

\rowcolor{gray!14}
\multicolumn{13}{@{}l}{\textit{Qwen3-8B}} \\

A-MEM
& \underline{74.60} & \underline{76.60}
& 68.70 & \underline{65.80}
& \underline{71.50} & \underline{68.20}
& 58.30 & \underline{62.50}
& \underline{76.50} & \underline{74.60}
& 68.50 & \underline{73.60} \\

MemZero
& 58.00 & 39.70 & 55.60 & 30.60
& 65.90 & 34.30 & 58.30 & 50.00
& 59.70 & 37.20 & 57.30 & 39.90 \\

FullContext
& \textbf{78.70} & \textbf{82.30}
& \textbf{85.90} & \textbf{89.10}
& 70.70 & \textbf{72.80}
& \textbf{100.00} & \textbf{100.00}
& \textbf{77.70} & \textbf{81.30}
& \textbf{80.30} & \textbf{83.10} \\

NaiveRAG
& 71.30 & 70.60
& \underline{69.70} & 63.10
& \underline{71.50} & 61.40
& 50.00 & 50.00
& 71.70 & 68.10
& \underline{69.60} & 69.00 \\

LightMem
& 61.30 & 42.40 & 59.60 & 41.90
& \textbf{76.40} & 47.90
& \underline{66.70} & 25.00
& 62.50 & 39.90 & 64.00 & 47.40 \\

\midrule
\addlinespace[2pt]

\rowcolor{gray!14}
\multicolumn{13}{@{}l}{\textit{Meditron3-Qwen2.5-14B}} \\

A-MEM
& 38.70 & 45.90 & 36.40 & 40.80
& 49.60 & \underline{60.10}
& \underline{66.70} & \underline{62.50}
& 41.30 & 44.70 & 38.40 & 51.60 \\

MemZero
& 43.20 & 28.60 & 47.50 & 23.50
& 52.00 & 36.90 & 25.00 & 25.00
& 43.00 & 24.70 & 46.90 & 36.10 \\

FullContext
& \textbf{61.90} & \textbf{65.80}
& \textbf{78.80} & \textbf{84.10}
& \textbf{63.40} & \textbf{64.30}
& \textbf{83.30} & \textbf{100.00}
& \textbf{64.00} & \textbf{66.50}
& \textbf{64.30} & \textbf{70.10} \\

NaiveRAG
& 50.10 & \underline{54.20}
& 39.40 & \underline{41.80}
& 50.40 & 56.20
& 41.70 & 41.70
& 49.70 & \underline{51.60}
& 47.70 & \underline{55.20} \\

LightMem
& \underline{50.70} & 39.70
& \underline{50.50} & 30.40
& \underline{54.50} & 40.20
& 58.30 & \underline{62.50}
& \underline{51.80} & 37.30
& \underline{50.40} & 42.00 \\

\midrule
\addlinespace[2pt]

\rowcolor{gray!14}
\multicolumn{13}{@{}l}{\textit{HuatuoGPT-3-27B}} \\

A-MEM
& \underline{91.90} & \underline{90.30}
& \underline{91.90} & \underline{86.00}
& \underline{95.90} & \underline{90.20}
& \underline{91.70} & 62.50
& \underline{93.80} & \underline{92.20}
& \underline{90.10} & \underline{85.20} \\

MemZero
& 83.10 & 58.30 & 85.90 & 53.90
& 91.10 & 59.30
& \textbf{100.00} & 66.70
& 84.50 & 58.60 & 84.80 & 57.40 \\

FullContext
& \textbf{95.68} & \textbf{95.98}
& \textbf{97.98} & \textbf{98.23}
& \textbf{95.93} & \textbf{95.99}
& 91.67 & \underline{87.50}
& \textbf{94.67} & \textbf{96.88}
& \textbf{97.87} & \textbf{94.88} \\

NaiveRAG
& 84.08 & 81.78 & 81.82 & 79.08
& 88.62 & 83.04
& 75.00 & \textbf{100.00}
& 85.00 & 82.00 & 83.20 & 81.71 \\

LightMem
& 81.40 & 57.20 & 84.80 & 50.10
& 93.50 & 51.60 & 66.70 & 50.00
& 83.00 & 55.30 & 83.20 & 56.20 \\

\midrule
\addlinespace[2pt]

\rowcolor{gray!14}
\multicolumn{13}{@{}l}{\textit{Qwen3-30B-A3B-Instruct-2507}} \\

A-MEM
& \underline{77.90} & \underline{81.40}
& \underline{73.70} & \underline{74.80}
& \textbf{78.90} & \textbf{82.00}
& \underline{91.70} & 62.50
& \textbf{81.20} & \textbf{83.90}
& 72.30 & \underline{75.10} \\

MemZero
& 64.10 & 41.90 & 64.60 & 28.00
& 71.50 & 36.60 & 41.70 & 50.00
& 65.00 & 37.50 & 64.50 & 43.90 \\

FullContext
& \textbf{80.30} & \textbf{84.10}
& \textbf{88.90} & \textbf{92.30}
& 74.00 & \underline{74.00}
& \textbf{100.00} & \textbf{100.00}
& \underline{80.00} & \underline{83.40}
& \textbf{81.60} & \textbf{84.60} \\

NaiveRAG
& 76.10 & 72.40 & 69.70 & 66.40
& \underline{78.00} & 67.90
& 50.00 & 50.00
& 77.00 & 72.40
& \underline{72.80} & 68.60 \\

LightMem
& 65.20 & 45.20 & 67.70 & 36.20
& \underline{78.00} & 39.80
& 58.30 & \underline{66.70}
& 68.00 & 42.80 & 65.30 & 45.50 \\

\bottomrule
\end{tabular}%
}
\end{table*}

\FloatBarrier

Table~\ref{tab:medical_memory_category} presents the performance of five memory methods across four language models on the RealMem medical subset, evaluating long-term conversational memory across four memory categories and two session types. Overall, FullContext achieves the strongest performance in most settings, particularly in dynamic updating and multi-session interactions. With Qwen3-30B, FullContext achieves QA and memory recall scores of 88.90 and 92.30 on dynamic updating, compared with 73.70 and 74.80 for A-MEM and 69.70 and 66.40 for NaiveRAG. This advantage is also observed with HuatuoGPT-3-27B, where FullContext achieves 97.98 in QA and 98.23 in recall.
Nevertheless, structured memory methods demonstrate selective advantages. In proactive alignment, A-MEM with Qwen3-30B achieves QA and recall scores of 78.90 and 82.00, outperforming FullContext by 4.90 and 8.00 points, respectively. LightMem also achieves a higher proactive alignment QA score than FullContext with Qwen3-8B, despite substantially lower memory recall. Notably, A-MEM exhibits reduced performance across multi-session interactions, with its QA score decreasing from 81.20 in single-session questions to 72.30 in multi-session questions under Qwen3-30B, whereas FullContext maintains comparable performance across both settings. 
\subsubsection{Effects of Conversation Length and Memory Distance on Memory Performance}

\begin{figure*}[t]
    \centering
    \includegraphics[width=\textwidth]{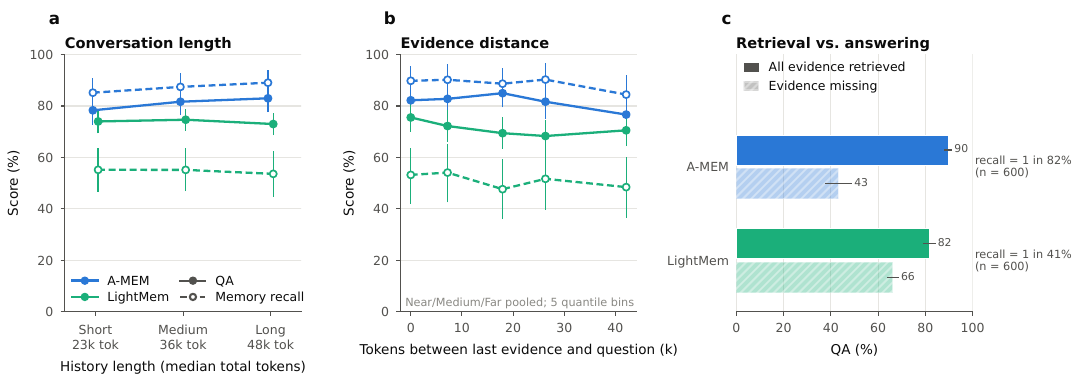}
    \caption{\textbf{Conversational memory robustness.}
    \textbf{a}, Effects of conversation history length on QA performance and memory recall under Short, Medium, and Long conditions.
    \textbf{b}, Relationship between evidence distance and memory performance, measured by the number of tokens separating the last relevant evidence from the question.
    \textbf{c}, QA performance conditioned on whether all required evidence was retrieved, comparing A-MEM and LightMem. Filled and open circles indicate complete and incomplete evidence retrieval, respectively.}
    \label{fig:realmem_length_distance}
\end{figure*}



We first investigated the sensitivity of long-term conversational memory to increasing history length using 100 questions from the RealMem medical subset (Fig.~\ref{fig:realmem_length_distance}a). Three experimental conditions were constructed by selectively removing irrelevant sessions preceding the earliest evidence-bearing session while preserving all required evidence and subsequent conversational content. This yielded Short, Medium, and Long histories with median lengths of 23,000, 36,000, and 48,000 tokens, respectively. Memory was independently reconstructed for each condition to evaluate the effects of history length on both retrieval and response generation.
Contrary to the expectation that longer histories would impair memory performance, increasing conversation length produced no consistent degradation across either memory system. With Qwen3-30B-A3B-Instruct-2507, A-MEM exhibited a modest increase in QA performance from 78.3 to 81.7 and 83.0 across the three conditions, accompanied by improvements in memory recall from 85.2 to 87.4 and 89.1. In contrast, LightMem maintained relatively stable QA scores of 74.0, 74.7, and 73.0, while memory recall remained substantially lower, ranging from 53.6 to 55.2. These observations suggest that, within the evaluated context lengths, additional irrelevant conversational history preceding the evidence does not necessarily compromise memory retrieval or downstream answering. The contrasting recall levels further indicate that differences in memory construction and retrieval mechanisms may be more influential than history length alone.

\subsubsection{Effects of Evidence Distance on Memory Retrieval and Utilization}

We examined whether the distance between relevant evidence and the current question affects memory performance (Fig.~\ref{fig:realmem_length_distance}b). Evidence distance was manipulated by repositioning irrelevant sessions while preserving the complete session set and total context length. Across five distance bins spanning up to 42,000 tokens, A-MEM maintained QA scores above 81\% at distances up to 26,000 tokens, declining to 76.7\% at the greatest distance. Memory recall showed a similar decline, whereas LightMem exhibited no consistent distance-dependent trend. Log-distance regression indicated modest negative associations with QA performance for both methods.
We further examined the relationship between evidence retrieval completeness and answering performance (Fig.~\ref{fig:realmem_length_distance}c). Across 600 question-condition observations, A-MEM retrieved all required evidence in 82\% of cases, achieving a QA score of 89.7 compared with 43.2 when retrieval was incomplete. LightMem achieved complete retrieval in 41\% of cases, with corresponding QA scores of 81.8 and 66.2. These results indicate a stronger association between evidence completeness and answering performance for A-MEM than for LightMem.
These findings suggest that conversation length and evidence distance do not consistently impair memory performance within the evaluated ranges. In contrast, complete evidence retrieval was strongly associated with successful answering, highlighting the importance of distinguishing memory retrieval quality from downstream evidence utilization.



\begin{figure*}[!t]
    \centering
    \includegraphics[width=\textwidth]{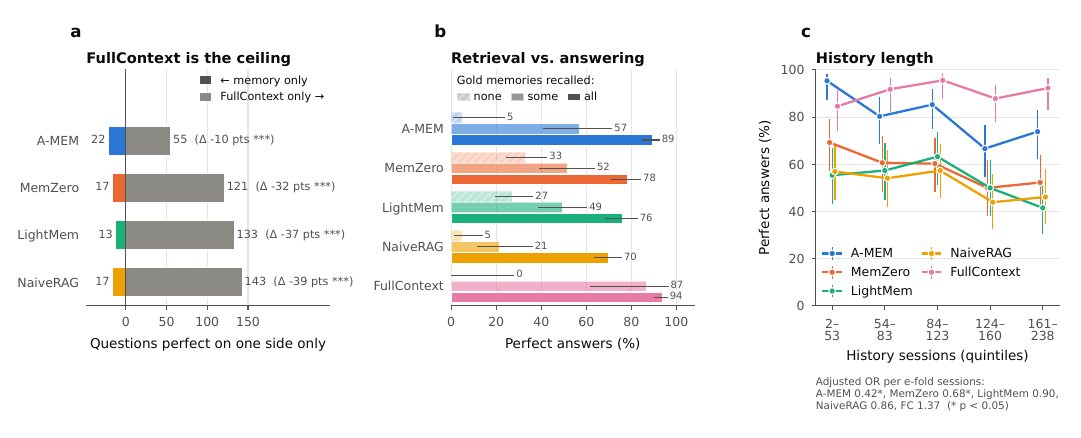}
    \caption{\textbf{Performance and robustness of long-term conversational memory systems.}
    \textbf{a}, Paired comparison of perfect-answer outcomes between four memory methods and FullContext (exact McNemar test; ***$P < 0.001$).
    \textbf{b}, Perfect-answer rates conditional on retrieving none, some, or all gold-standard memories.
    \textbf{c}, Perfect-answer rates across five conversation history-length groups, with adjusted odds ratios for history length.
    Error bars indicate 95\% confidence intervals.}
    \label{fig:realmem_main}
\end{figure*}

\subsubsection{Association Between Conversation History Length and Memory Performance}

We examined the relationship between conversation history length and question-answering performance across all RealMem-Med test questions (Fig.~\ref{fig:realmem_main}c). Questions were divided into five approximately equal-sized groups according to the number of historical conversation sessions, ranging from 2 to 238 sessions. A-MEM showed a decline in perfect-answer accuracy from approximately 95\% in the shortest-history group to 74\% in the longest-history group. MemZero exhibited a similar downward trend, whereas LightMem and NaiveRAG maintained lower accuracy with more variable patterns. In contrast, FullContext consistently achieved high accuracy across history-length groups.
Regression analysis further identified significant negative associations between history length and answering accuracy for A-MEM (adjusted odds ratio [OR] = 0.42, $p<0.05$) and MemZero (OR = 0.68, $p<0.05$). No statistically significant associations were observed for LightMem, NaiveRAG, or FullContext. These results suggest that longer conversation histories are associated with reduced answering performance for certain memory systems. However, because the analysis compares different questions with naturally varying history lengths, the observed relationships may also reflect differences in question complexity and information requirements rather than a direct effect of history length.

\subsubsection{Evaluating Retrieval and Answering in Long-Term Conversational Memory}

We further evaluated the effectiveness of long-term conversational memory using 325 questions from the medical subset of RealMem. We compared four memory-based approaches, A-MEM, MemZero, LightMem, and NaiveRAG, against FullContext, which provides the complete conversation history to the language model. Beyond overall answer accuracy, we examined three complementary aspects of memory performance: question-level differences relative to FullContext, the relationship between evidence retrieval completeness and answer correctness, and the association between conversation history length and answering performance (Fig.~\ref{fig:realmem_main}).

\paragraph{Performance gap relative to full conversation history.}
FullContext consistently outperformed all four memory-based approaches (Fig.~\ref{fig:realmem_main}a). In paired question-level comparisons, FullContext answered 55 questions correctly that A-MEM failed to answer, whereas A-MEM uniquely answered 22 questions correctly, corresponding to a 10-percentage-point difference in overall perfect-answer rate. Larger gaps were observed for MemZero, LightMem, and NaiveRAG, with differences of 32, 37, and 39 percentage points, respectively. All four paired comparisons were significant under exact McNemar tests, with $P < 0.001$. These results indicate that existing memory-based approaches do not fully preserve the answering performance achievable when the complete conversation history remains accessible.

\paragraph{Evidence retrieval completeness and answer correctness.}
To examine the relationship between memory retrieval and downstream answering, we stratified questions according to whether none, some, or all gold-standard memories were retrieved (Fig.~\ref{fig:realmem_main}b). Across memory methods, perfect-answer rates generally increased with retrieval completeness, although substantial differences remained in how effectively retrieved evidence was utilized. A-MEM showed a pronounced increase in accuracy, from 5\% when no gold-standard memories were retrieved to 57\% with partial retrieval and 89\% with complete retrieval. Similar patterns were observed for MemZero and LightMem, which achieved perfect-answer rates of 78\% and 76\%, respectively, when all required memories were retrieved. In contrast, NaiveRAG achieved only 70\% despite complete retrieval, highlighting limitations in downstream evidence utilization. FullContext achieved higher accuracy, reaching 94\% when all required evidence was available. Notably, MemZero and LightMem correctly answered 33\% and 27\% of questions, respectively, even when no gold-standard memories were retrieved. Together, these findings demonstrate that retrieval completeness is strongly associated with answering performance, but access to relevant evidence alone does not guarantee correct utilization. Because the retrieval categories contain different questions, these associations do not establish a causal effect of retrieval completeness on answer accuracy.

\paragraph{Association between history length and memory performance.}
We additionally examined performance across five groups defined by the number of historical conversation sessions, ranging from 2 to 238 sessions (Fig.~\ref{fig:realmem_main}c). A-MEM achieved a perfect-answer rate of approximately 95 per cent among questions with the shortest histories, compared with 67--74 per cent among questions with longer histories. MemZero, LightMem, and NaiveRAG exhibited similar descriptive declines, whereas FullContext maintained comparatively stable performance across history-length groups. After adjusting for question category, evidence count, and session type, history length remained negatively associated with perfect-answer outcomes for A-MEM and MemZero, with adjusted odds ratios of 0.42 and 0.68, respectively. No statistically significant associations were observed for LightMem, NaiveRAG, or FullContext.

\subsection{Performance Evaluation of HMRU}

\subsubsection{Improving Biomedical Memory Retrieval and Utilization}

We evaluated the proposed HMRU framework across six biomedical tasks (Table~\ref{tab:biomem_prompt_adapter}). We selected A-MEM as the baseline memory method and HuatuoGPT-3-27B as the frozen language model backbone, enabling a controlled comparison of retrieval and memory utilization strategies without modifying the underlying model parameters. Among the evaluated configurations, HMRU-I, which integrates hybrid retrieval, cross-encoder reranking, and task-aware memory utilization instructions, achieved the strongest overall performance, attaining the highest score in 15 of 24 evaluation metrics, including two ties. The largest improvements were observed in biomedical question answering. Compared with the A-MEM baseline, HMRU-I increased exact-match F1 from 39.27 to 43.48 on RiTeK-PharmKG and from 52.94 to 57.64 on RiTeK-ADint, corresponding to gains of 4.21 and 4.70 percentage points, respectively. HMRU-I also outperformed retrieval enhancement alone, increasing exact-match F1 from 42.95 to 43.48 on PharmKG and from 54.60 to 57.64 on ADint. These results suggest that task-aware memory utilization provides complementary benefits beyond hybrid retrieval and reranking.
The effects were more heterogeneous across clinical prediction tasks. HMRU-C, which additionally incorporates compact memory representations, achieved the highest mortality prediction AUPRC of 10.03 and AUROC of 61.87, whereas HMRU-I obtained the highest readmission AUPRC of 62.17. In contrast, neither configuration consistently improved opioid overdose prediction, for which the A-MEM baseline retained the highest AUPRC of 4.94. Collectively, these findings demonstrate that hybrid retrieval and task-aware memory utilization improve biomedical question-answering performance, whereas their effectiveness varies across clinical prediction tasks.

\begin{table*}[!t]
\centering
\scriptsize
\renewcommand{\arraystretch}{1.15}
\setlength{\tabcolsep}{3pt}

\resizebox{\textwidth}{!}{%
\begin{tabular}{@{}l*{20}{r}@{}}
\toprule

\multirow{2}{*}{\textbf{Method}}
& \multicolumn{6}{c}{\textbf{RiTeK-PharmKG}}
& \multicolumn{6}{c}{\textbf{RiTeK-ADint}}
& \multicolumn{2}{c}{\textbf{Mort.-MIMIC-III}}
& \multicolumn{2}{c}{\textbf{Readm.-MIMIC-III}}
& \multicolumn{2}{c}{\textbf{TCGA-Reports}}
& \multicolumn{2}{c}{\textbf{MIMIC-IV OOD}} \\

\cmidrule(lr){2-7}
\cmidrule(lr){8-13}
\cmidrule(lr){14-15}
\cmidrule(lr){16-17}
\cmidrule(lr){18-19}
\cmidrule(lr){20-21}

& $P^*$ & $R^*$ & $F_1^*$
& $P^\dagger$ & $R^\dagger$ & $F_1^\dagger$
& $P^*$ & $R^*$ & $F_1^*$
& $P^\dagger$ & $R^\dagger$ & $F_1^\dagger$
& AUPRC & AUROC
& AUPRC & AUROC
& AUPRC & AUROC
& AUPRC & AUROC \\

\midrule

Base method
& 39.95 & 39.23 & 39.27
& 42.18 & 41.40 & 41.51
& 53.32 & 53.08 & 52.94
& 58.84 & 58.62 & 58.53
& \underline{9.81} & 59.88
& \underline{61.91} & 56.69
& \textbf{99.06} & \underline{99.93}
& \textbf{4.94} & \underline{82.16} \\

HMRU-R
& 43.47 & 43.04 & 42.95
& 45.53 & \textbf{45.05} & 45.01
& 55.08 & 54.53 & 54.60
& 60.28 & 59.86 & 59.90
& 9.66 & 60.43
& 61.62 & 57.68
& \underline{98.84} & \textbf{99.94}
& \underline{4.28} & \textbf{84.08} \\

HMRU-I
& \textbf{44.31} & \textbf{43.29} & \textbf{43.48}
& \textbf{46.01} & \underline{45.02} & \textbf{45.23}
& \textbf{58.30} & \textbf{57.33} & \textbf{57.64}
& \textbf{63.34} & \textbf{62.57} & \textbf{62.81}
& 9.67 & \underline{61.16}
& \textbf{62.17} & \underline{57.89}
& \underline{98.84} & \textbf{99.94}
& 4.17 & 78.43 \\

HMRU-C
& \underline{44.17} & \underline{43.26} & \underline{43.35}
& \underline{45.92} & \textbf{45.05} & \underline{45.16}
& \underline{56.54} & \underline{55.67} & \underline{55.91}
& \underline{62.00} & \underline{61.29} & \underline{61.50}
& \textbf{10.03} & \textbf{61.87}
& 61.72 & \textbf{57.92}
& \underline{98.84} & \textbf{99.94}
& 3.91 & 76.96 \\

\bottomrule
\end{tabular}%
}

\caption{
Comparison of HMRU variants and baseline memory performance across six biomedical tasks.
All scores are percentages. Bold and underlined values indicate the best and second-best distinct scores for each metric, respectively.
}
\label{tab:biomem_prompt_adapter}
\end{table*}

\FloatBarrier

\begin{table*}[!t]
\centering
\scriptsize
\renewcommand{\arraystretch}{1.15}
\setlength{\tabcolsep}{4pt}

\resizebox{0.8\textwidth}{!}{%
\begin{tabular}{@{}l*{12}{r}@{}}
\toprule

\multirow{3}{*}{\textbf{Method}}
& \multicolumn{8}{c}{\textbf{Memory Category}}
& \multicolumn{4}{c}{\textbf{Session Type}} \\

\cmidrule(lr){2-9}
\cmidrule(lr){10-13}

& \multicolumn{2}{c}{\textbf{Dyn. Inc.}}
& \multicolumn{2}{c}{\textbf{Dyn. Upd.}}
& \multicolumn{2}{c}{\textbf{Proac. Align.}}
& \multicolumn{2}{c}{\textbf{Temp. Reas.}}
& \multicolumn{2}{c}{\textbf{Single Sess.}}
& \multicolumn{2}{c}{\textbf{Multi Sess.}} \\

\cmidrule(lr){2-3}
\cmidrule(lr){4-5}
\cmidrule(lr){6-7}
\cmidrule(lr){8-9}
\cmidrule(lr){10-11}
\cmidrule(lr){12-13}

& QA & Rec.
& QA & Rec.
& QA & Rec.
& QA & Rec.
& QA & Rec.
& QA & Rec. \\

\midrule


Base Method
& 91.90 & 90.34
& \underline{91.92} & 86.01
& \textbf{95.93} & 90.16
& \textbf{91.67} & \underline{62.50}
& \textbf{93.83} & 92.25
& 90.13 & 85.18 \\

HMRU-R
& \textbf{93.25} & 92.63
& \textbf{94.95} & \textbf{91.31}
& \underline{95.12} & \underline{91.99}
& \textbf{91.67} & \textbf{75.00}
& \textbf{93.83} & \textbf{93.88}
& \textbf{93.33} & \underline{89.51} \\

HMRU-I
& \underline{93.12} & \underline{92.73}
& \underline{91.92} & \underline{86.26}
& \underline{95.12} & \textbf{96.86}
& \textbf{91.67} & \textbf{75.00}
& \underline{93.50} & \underline{93.62}
& \underline{92.80} & \textbf{90.39} \\

HMRU-T
& 90.96 & \textbf{92.87}
& 79.80 & 83.33
& 88.62 & 91.38
& \underline{66.67} & \textbf{75.00}
& 90.33 & 93.38
& 87.47 & 88.49 \\

\bottomrule
\end{tabular}%
}

\caption{
Comparison of HMRU variants and baseline memory performance on RealMem using HuatuoGPT-3-27B.
QA and Rec. denote QA Score and Memory Recall, respectively.
All scores are percentages. Bold and underlined values indicate the best and second-best distinct scores for each metric, respectively.
}
\label{tab:realmem_bm25_prompt_adapter}
\end{table*}

\FloatBarrier

\begin{figure*}[!t]
    \centering
    \includegraphics[width=\textwidth]{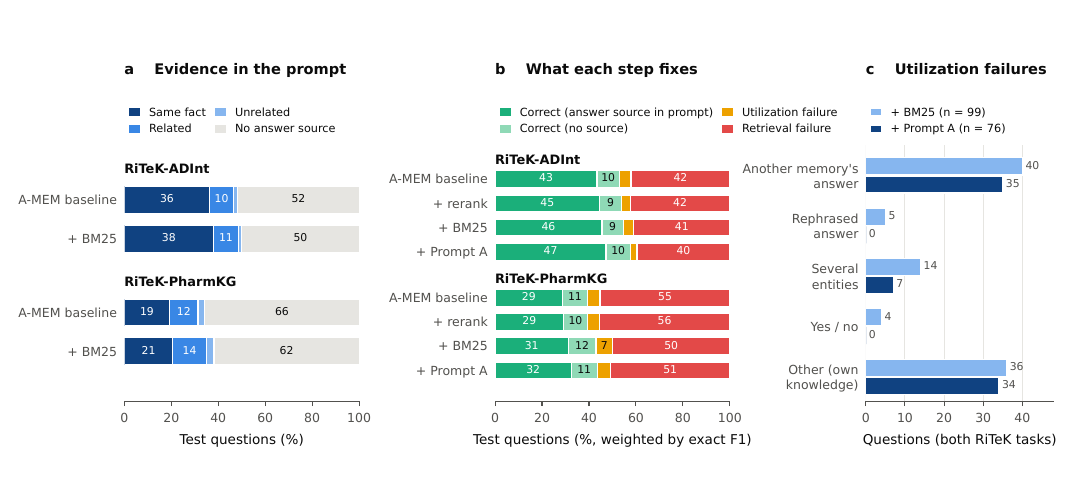}
    \caption{
   Mechanistic analysis of Hybrid Memory Retrieval and Utilization on biomedical question answering.
    \textbf{a}, Distribution of retrieved evidence categories before and after BM25 augmentation.
    \textbf{b}, Decomposition of answer performance and failure modes across successive retrieval and utilization enhancements.
    \textbf{c}, Distribution of evidence utilization errors before and after task-aware adaptation.
    Results are reported for RiTeK-PharmKG and RiTeK-ADInt using HuatuoGPT-3-27B.
    }
    \label{fig:hmru_mechanism}
\end{figure*}

\subsubsection{Mechanistic Analysis of HMRU Improvements.}
To investigate the sources of performance gains, we analyzed evidence availability and utilization across successive HMRU components on RiTeK-PharmKG and RiTeK-ADInt (Fig.~\ref{fig:hmru_mechanism}). Hybrid retrieval increased the proportion of questions with answer-bearing memories, reducing the absence of answer sources from 65.8 to 61.8 per cent on PharmKG and from 51.7 to 50.0 per cent on ADInt (Fig.~\ref{fig:hmru_mechanism}a). However, the contributions of retrieval enhancement and task-aware utilization differed between tasks (Fig.~\ref{fig:hmru_mechanism}b). On PharmKG, BM25 augmentation reduced retrieval failures from 55.3 to 50.0 per cent and increased exact-match F1 from 39.3 to 43.0. In contrast, on ADInt, the largest incremental improvement followed the introduction of the Task-Aware Memory Utilization Adapter (TAMUA), which increased exact-match F1 from 54.6 to 57.6 while reducing utilization failures from 4.5 to 2.8 per cent. Further analysis showed that TAMUA reduced the combined number of utilization failures across both tasks from 99 to 76, eliminating observed entity rephrasing and inappropriate yes/no responses while reducing errors involving multiple entities from 14 to 7 (Fig.~\ref{fig:hmru_mechanism}c). Nevertheless, incorrect selection of answers from competing memories remained a frequent source of error. Together, these findings suggest that hybrid retrieval and task-aware memory utilization address complementary limitations, with their relative contributions depending on the characteristics of the biomedical question-answering task.

\subsubsection{Improving Long-Term Conversational Memory Utilization}

We further evaluated HMRU on the RealMem medical subset using A-MEM as the baseline memory system and HuatuoGPT-3-27B as the frozen language model (Table~\ref{tab:realmem_bm25_prompt_adapter}). We investigated three configurations: HMRU-R, which combines hybrid retrieval with cross-encoder reranking; HMRU-I, which additionally incorporates task-aware instructions for conversational continuity, preference preservation, and temporal consistency; and HMRU-T, which reorganizes the same retrieved memories into chronological conversation sequences and introduces structured state tracking of the ongoing dialogue, user profile, and temporal updates. HMRU-R achieved the strongest overall performance, improving QA scores from 91.90 to 93.25 on Dynamic Incremental, from 91.92 to 94.95 on Dynamic Updating, and from 90.13 to 93.33 on multi-session questions. Corresponding memory recall also improved, including an increase from 86.01 to 91.31 on Dynamic Updating. Adding task-aware utilization instructions yielded further improvements in specific memory capabilities. HMRU-I increased Proactive Alignment recall from 90.16 to 96.86 and multi-session recall from 85.18 to 90.39, exceeding both the baseline and HMRU-R. However, chronological memory organization and explicit state tracking did not provide consistent additional benefits. HMRU-T showed substantial reductions in Dynamic Updating QA, from 94.95 under HMRU-R to 79.80, and Temporal Reasoning QA, from 91.67 to 66.67. These findings suggest that hybrid retrieval provides broad improvements in conversational memory performance, whereas task-aware utilization instructions offer selective benefits for preference-sensitive and multi-session tasks. More extensive temporal restructuring does not necessarily improve memory utilization, highlighting the importance of preserving retrieval relevance alongside conversational context.

\begin{figure*}[!t]
    \centering
    \includegraphics[width=\textwidth]{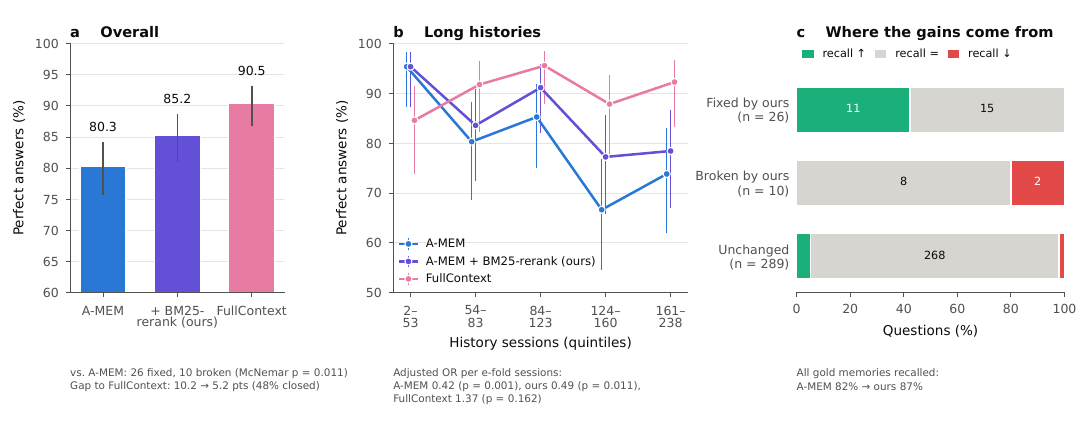}
    \caption{
    Performance improvements of HMRU-R on long-term conversational memory using HuatuoGPT-3-27B.
    \textbf{a}, Overall perfect-answer rates compared with the baseline and FullContext.
    \textbf{b}, Performance across five conversation history-length groups.
    \textbf{c}, Question-level improvements and regressions stratified by changes in gold-memory recall.
    }
    \label{fig:realmem_ours}
\end{figure*}

\subsubsection{Improving Long-Term Conversational Memory through Hybrid Retrieval}

We further examined the effectiveness of HMRU-R on RealMem-mem, using A-MEM as the baseline memory system and HuatuoGPT-3-27B as the frozen language model (Fig.~\ref{fig:realmem_ours}). HMRU-R increased the overall perfect-answer rate from 80.3 to 85.2 per cent, reducing the performance gap relative to FullContext from 10.2 to 5.2 percentage points, corresponding to a 48 per cent reduction in the original gap (Fig.~\ref{fig:realmem_ours}a). Paired question-level analysis identified 26 questions corrected by HMRU-R and 10 previously correct questions that became incorrect, yielding a statistically significant improvement under the exact McNemar test, with $P=0.011$.
We next assessed whether these improvements extended to longer conversation histories (Fig.~\ref{fig:realmem_ours}b). HMRU-R generally outperformed the baseline across history-length groups, with particularly pronounced gains in longer conversations. Nevertheless, after adjustment for question category, evidence count, and session type, history length remained negatively associated with perfect-answer outcomes for both A-MEM and HMRU-R, with adjusted odds ratios of 0.42 and 0.49, respectively. Thus, hybrid retrieval improved overall answering performance but did not eliminate the association between longer histories and reduced accuracy.
To further characterize the sources of improvement, we examined changes in gold-memory recall among questions with altered answering outcomes (Fig.~\ref{fig:realmem_ours}c). Among the 26 questions corrected by HMRU-R, 11 exhibited increased recall of required evidence, whereas 15 were corrected without a change in gold-memory recall. Conversely, among the 10 questions that became incorrect, 8 showed unchanged recall and 2 showed reduced recall. Overall, the proportion of questions for which all required memories were retrieved increased from 82 to 87 per cent. These findings indicate that HMRU-R improves both evidence accessibility and downstream answering, while the substantial number of corrections without recall changes suggests that retrieval ranking and evidence presentation may also contribute to its effectiveness.

\section{Discussion}

\paragraph{Biomedical memory performance depends on task characteristics}
This study examined how agentic memory systems construct, retrieve, update and use information across biomedical knowledge reasoning, clinical prediction and long-term conversational tasks. Across six biomedical benchmarks, four language models and five memory approaches, no method consistently outperformed the others. The analyses instead identified three recurring limitations: relevant evidence may not be retrieved, retrieved evidence may be used incorrectly, and evolving memory representations may diverge from their source records. We further investigated these issues in long-term conversation and developed Hybrid Memory Retrieval and Utilization (HMRU), a training-free framework that improves retrieval and evidence use without updating language-model parameters. These findings suggest that memory quality cannot be characterized by storage capacity or retrieval volume alone.
The relative advantages of memory methods changed with both the task and the language-model backbone (Table~\ref{tab:training_data_memory}). With HuatuoGPT-3-27B, A-MEM achieved exact-match F1 scores of 39.27 on RiTeK-PharmKG and 52.94 on RiTeK-ADInt, compared with 39.04 and 51.63 for NaiveRAG. With Qwen3-30B, however, NaiveRAG outperformed A-MEM on both tasks. The differences were more substantial in clinical prediction: with HuatuoGPT-3-27B, NaiveRAG achieved an opioid overdose AUPRC of 15.61, compared with 4.94 for A-MEM, whereas MemZero performed best on mortality prediction with Meditron3-Qwen2.5-14B. Thus, more elaborate memory organization does not necessarily produce better downstream results. Increasing the memory construction set from 250 to 1,000 records improved A-MEM F1 on PharmKG from 27.66 to 39.27, but larger memory collections, deeper retrieval and class rebalancing did not uniformly benefit clinical prediction.
These task differences may reflect distinct relationships between semantic similarity and the information needed for prediction. In TCGA-Reports, retrieved examples were strongly enriched for the same cancer category, with retrieval lift exceeding 15 (Fig.~\ref{fig:memory_layers}f). By contrast, label enrichment was weaker and less consistent for mortality and readmission. Two patients with similar diagnoses or clinical histories may nevertheless experience different future outcomes. Consequently, retrieval objectives that work well for identifying explicit biomedical facts or diagnostic categories may be insufficient for prospective risk prediction, which requires attention to longitudinal patterns and clinically meaningful differences between patients.
Outcome prediction, however, requires more than identifying a superficially similar patient. The relevance of historical diagnoses depends on their timing, combinations, severity and relation to the target event. A memory store assembled from examples with matching labels could also overrepresent prevalent outcomes without providing meaningful discrimination. This is particularly important when positive cases are uncommon, as in opioid overdose prediction, where AUPRC and retrieval enrichment should be interpreted relative to event prevalence. Our findings therefore argue against using a single retrieval objective across biomedical applications. A useful next step would compare semantically nearest neighbors with temporally aligned, clinically matched and outcome-aware candidates under the same retrieval budget, while preserving a separate held-out evaluation cohort.

\paragraph{Evidence availability and evidence utilization are distinct limitations}
Our evidence-level analyses help explain why retrieving additional memories does not always improve question answering. Across the RiTeK datasets, the presence of an answer-matching source was associated with substantially greater accuracy: retrieval-based methods generally answered 80--88\% of questions correctly when such a source was available, compared with 12--16\% when it was absent (Fig.~\ref{fig:memory_layers}c). Yet an answer-matching record did not necessarily support the same biomedical fact. On RiTeK-ADInt, A-MEM, LightMem and NaiveRAG each retrieved same-fact sources for approximately 35\% of questions, without a consistent coverage advantage for the more structured methods (Fig.~\ref{fig:memory_layers}b). Accuracy also increased with the semantic similarity between the question and its answer-matching source, approaching 100\% in the highest similarity bin (Fig.~\ref{fig:memory_layers}e). These observations underscore the difference between recovering an answer entity and retrieving evidence that addresses the correct biomedical relation.
Evidence utilization introduced a separate challenge. Incorrect predictions exactly matched answers contained in retrieved records for 16--21\% of RiTeK-ADInt questions and 24--29\% of RiTeK-PharmKG questions across the retrieval-based approaches (Fig.~\ref{fig:memory_layers}d). Such overlap is consistent with interference from competing answers, although it does not prove that the model copied a specific memory. Similarly, even complete retrieval of annotated evidence did not always produce correct answers in the long-term conversational evaluation. Thus, memory systems must both recover appropriate evidence and distinguish among potentially conflicting sources during generation. These analyses are observational: questions with retrievable evidence may be inherently easier, and causal effects require interventions that remove, reorder or replace retrieved memories. Nevertheless, they support evaluating factual coverage and downstream evidence use as separate properties of a memory system.

\paragraph{Iterative memory evolution does not guarantee source fidelity}

A further concern is whether generated memory representations faithfully preserve their original information. Agentic memory mechanisms may summarize incoming records and rewrite existing notes as new information arrives. Although these operations can improve organization, they can also introduce unsupported contextual statements. In our A-MEM analysis, the proportion of memories with context not supported by their original training records increased from 12\% to 44\% on MIMIC-IV opioid overdose prediction and from 16\% to 35\% on RiTeK-PharmKG (Fig.~\ref{fig:amem_evolution}c). In contrast, unsupported-context prevalence remained below 2\% in both MIMIC-III tasks, indicating substantial task dependence.
Repeated rewriting was associated with the same pattern. In MIMIC-IV OOD, approximately 10\% of memories without rewrites contained unsupported context, compared with 80\% among those rewritten at least five times (Fig.~\ref{fig:amem_evolution}b). Longitudinal examination of frequently rewritten memories revealed extended sequences of unsupported versions, but some notes also returned to source-supported states (Fig.~\ref{fig:amem_evolution}d). These findings do not establish that rewriting causes factual errors: complex records may be rewritten more frequently, and information unsupported by a particular source is not necessarily false. They nevertheless challenge the assumption that iterative refinement invariably improves memory quality. Source-linked provenance, verification against original records and explicit separation of source facts from generated interpretations merit investigation, especially when updated notes are reused in subsequent reasoning.

\paragraph{Long-term conversational memory requires accessible and complete evidence}
The RealMem experiments revealed a substantial gap between structured memory and access to the full conversation history. On 325 medical-subset questions, FullContext outperformed all four memory-based approaches in overall perfect-answer accuracy (Fig.~\ref{fig:realmem_main}a). Relative to A-MEM, it uniquely answered 55 questions correctly, whereas A-MEM uniquely answered 22; the paired difference was significant under the exact McNemar test ($P<0.001$). This is not evidence that FullContext is correct on every question, but it indicates that existing memory systems lose information or fail to exploit information that remains accessible in the complete history. Evidence completeness also mattered: A-MEM's perfect-answer rate increased from approximately 5\% when no gold memories were retrieved to 89\% when all were retrieved (Fig.~\ref{fig:realmem_main}b). However, some errors persisted despite complete evidence retrieval, reinforcing the distinction between access and utilization.
Conversation length produced a more nuanced picture. Across naturally occurring histories containing 2--238 sessions, longer histories were negatively associated with answering performance for A-MEM (adjusted odds ratio 0.42) and MemZero (0.68), but not significantly for the other methods (Fig.~\ref{fig:realmem_main}c). In contrast, when the same 100 questions were evaluated under controlled histories with median lengths of 23,000, 36,000 and 48,000 tokens, neither A-MEM nor LightMem exhibited a consistent performance decline (Fig.~\ref{fig:realmem_length_distance}a). Relocating relevant evidence within a fixed history also produced limited overall deterioration, although A-MEM declined at the greatest tested distance (Fig.~\ref{fig:realmem_length_distance}b). Across 600 question-condition observations, A-MEM achieved a QA score of 89.7 with complete evidence retrieval versus 43.2 with incomplete retrieval (Fig.~\ref{fig:realmem_length_distance}c). These results suggest that natural conversation length is entangled with evidence distribution, updating requirements and question complexity; token count alone is not an adequate explanation for long-term memory failures.
The contrast between observational and controlled results is methodologically important. Longer natural histories may contain more factual revisions, references distributed across sessions, or multiple related entities that must be reconciled. Removing irrelevant material from a fixed question, by comparison, does not necessarily reproduce these challenges. Thus, the absence of deterioration in a controlled length experiment should not be interpreted as proof that memory remains robust in arbitrarily long real-world interactions. Benchmarks should independently vary history length, number of required evidence items, temporal update frequency and the placement of contradictory or superseded statements.

\paragraph{Hybrid retrieval and task-aware utilization offer complementary benefits}
HMRU was designed to address the retrieval and utilization limitations identified above. It combines lexical and semantic retrieval with cross-encoder reranking and task-aware instructions, while retaining the underlying memory bank and frozen language model. On the two biomedical question-answering tasks, HMRU-I improved exact-match F1 from 39.27 to 43.48 on PharmKG and from 52.94 to 57.64 on ADInt relative to A-MEM (Table~\ref{tab:biomem_prompt_adapter}). The component analysis indicated different sources of improvement. On PharmKG, BM25 augmentation reduced classified retrieval failures from 55.3\% to 50.0\%; on ADInt, task-aware instructions improved F1 from 54.6 to 57.6 after retrieval enhancement (Fig.~\ref{fig:hmru_mechanism}). Across both tasks, identified utilization failures decreased from 99 to 76 following prompt adaptation. These results support combining evidence acquisition with better interpretation rather than treating retrieval alone as sufficient.
The benefits were not universal. HMRU-C modestly improved mortality AUPRC from 9.81 to 10.03, and HMRU-I increased readmission AUPRC from 61.91 to 62.17, but neither configuration improved opioid overdose AUPRC over the A-MEM baseline. This limitation is consistent with the weaker same-label enrichment observed in clinical prediction: identifying relevant medical information and discriminating future outcomes are not equivalent objectives. Clinical prediction may require outcome-aware retrieval and uncertainty-sensitive utilization strategies beyond those effective for factual question answering.
In long-term conversation, HMRU-R increased the RealMem perfect-answer rate from 80.3\% to 85.2\%, reducing the gap to FullContext (90.5\%) by nearly half (Fig.~\ref{fig:realmem_ours}). Paired comparison identified 26 corrected questions and 10 newly incorrect answers (exact McNemar $P=0.011$). Gold-memory recall was unchanged in 15 of the 26 corrected cases, suggesting that evidence ranking or presentation may contribute beyond retrieving additional annotated memories. More extensive restructuring was not consistently beneficial: chronological HMRU-T reduced accuracy on some dynamic updating and temporal reasoning questions. Thus, utilization strategies should reflect the specific information requirements of the task, rather than assuming that additional structure always helps.

\section{Methods}

\subsection{Datasets, Memory Baselines, and Language Models}
\label{sec:datasets_baselines_llms}

\paragraph{Datasets.}
We evaluated biomedical agentic memory on six tasks covering knowledge reasoning, clinical outcome prediction, and pathology classification. RiTeK-PharmKG~\citep{huang2026ritek} and RiTeK-ADInt~\citep{huang2026ritek} assess biomedical question answering over relational knowledge. MIMIC-III Mortality and MIMIC-III Readmission are clinical outcome prediction tasks derived from MIMIC-III~\citep{johnson2016mimic}. TCGA-Reports supports cancer classification from pathology report text~\citep{kefeli2024tcreports}, whereas MIMIC-IV OOD ~\citep{li2026overdosemoe} evaluates opioid overdose prediction using electronic health records from MIMIC-IV. We further evaluated long-term conversational memory from the medical subset of RealMem~\citep{bian2026realmem}, which includes multi-session interactions requiring information retention, updating, and temporal reasoning. We report exact-match and ROUGE-1 precision, recall, and F1 for the RiTeK tasks; AUPRC and AUROC are used for the four classification and prediction tasks. For the RealMem evaluation, we used GPT-4o-mini as an LLM-based judge to assess the correctness of generated answers. 

\paragraph{Memory baselines with Language model backbones.}
We compared five memory configurations spanning structured memory management, direct retrieval, and full-context access. A-MEM\citep{xu2026mem} constructs linked memory notes and supports dynamic indexing and iterative memory evolution. LightMem~\citep{fang2026lightmem} uses lightweight filtering and memory consolidation to manage persistent information. MemZero was evaluated as an additional persistent-memory approach, with Mem0~\citep{chhikara2025mem0} providing the related framework for memory extraction, retrieval, and updating. NaiveRAG~\citep{lewis2020retrieval} serves as an embedding-based retrieval baseline that selects relevant source records without evolving a structured memory bank, following the general retrieval-augmented generation paradigm. FullContext provides the available source records or conversation history directly to the language model without selective retrieval. The same underlying training records and evaluation questions were used for method comparisons within each task.  We evaluated four open-weight language models: Qwen3-8B\footnote{https://huggingface.co/Qwen/Qwen3-8B}, Qwen3-30B-A3B-Instruct-2507\footnote{https://huggingface.co/Qwen/Qwen3-30B-A3B-Instruct-2507}, Meditron3-Qwen2.5-14B\footnote{https://huggingface.co/EPFLiGHT/Meditron3-Qwen2.5-14B}, and HuatuoGPT-3-27B\footnote{https://huggingface.co/FreedomIntelligence/HuatuoGPT-3-27B}.

\subsection{Memory Retrieval Quality and Utilization}
\paragraph{Analysis design.}
To examine why different memory systems yield different downstream outcomes, we analyzed test-set retrieval and prediction traces from A-MEM, MemZero, LightMem, NaiveRAG, and FullContext across six biomedical benchmarks. The analyzed predictions were generated by HuatuoGPT-3-27B, with each method drawing on the same 1,000 training examples per dataset for memory construction. Only memory items that actually entered the final inference prompt were included. We organized the analysis into three layers: answer or label exposure, answer retention following memory construction, and relevance of the retrieved evidence. The full three-layer analysis was applied to RiTeK-ADInt and RiTeK-PharmKG. For TCGA-Reports, MIMIC-IV OOD, MIMIC-III Mortality, and MIMIC-III Readmission, we assessed label exposure and source relevance, but did not apply literal answer-retention checks because class labels were not consistently preserved as explicit text in the memory representations.

For the two RiTeK datasets, let $q_i$, $y_i$, and $\hat y_i$ denote the test question, reference answer, and generated answer. Let $\mathcal M_i$ contain the memory entries included in its final prompt, and let $\mathcal S(m)$ denote the original training records associated with memory entry $m$. When one memory entry merged several records, all recorded source identifiers were used for provenance tracing. We defined answer-source exposure as
\begin{equation}
A_i=\mathbb{1}\!\left[\exists m\in\mathcal M_i,\;s\in\mathcal S(m):
\operatorname{norm}(y_s)=\operatorname{norm}(y_i)\right],
\label{eq:memory_answer_exposure}
\end{equation}
where $y_s$ is the original training answer and $\operatorname{norm}$ denotes case, punctuation, and whitespace normalization. This criterion identifies sources sharing the answer, not necessarily sources supporting the same biomedical fact. Conditional on $A_i=1$, we further checked whether the normalized reference answer appeared as a whole-word sequence in the corresponding stored memory text. This second layer assesses explicit answer retention in \emph{retrieved} memory, rather than information retention across the entire memory store.

To identify factual relevance, an independent Qwen3-30B-A3B-Instruct-2507 evaluator classified each answer-matching test--source pair as \emph{same fact}, \emph{related}, or \emph{unrelated}, using the test question and answer alongside the source question and answer. Evaluation used temperature zero; source pairs were deduplicated across methods. Each test question was assigned its strongest retrieved evidence category, in the order same fact, related, and unrelated, or \emph{no answer source} when $A_i=0$. Separately, BAAI/bge-m3 embeddings with a maximum input length of 512 tokens were used to compute cosine similarities between test and source questions. For TCGA-Reports, the same embedding model measured pathology-report similarity. For the three binary clinical tasks, similarity was computed as Jaccard overlap between normalized sets of diagnosis codes without using an additional LLM judge.

\paragraph{Layer 1: Answer exposure and label consistency.}
For RiTeK-ADInt and RiTeK-PharmKG, we traced retrieved memories to their original training records using source identifiers and compared the corresponding reference answers with the test ground truth after text normalization. We measured the proportion of questions with at least one answer-matching source and the prediction accuracy conditioned on its presence or absence. For TCGA-Reports, we calculated the proportion of retrieved training reports sharing the correct cancer label, nearest-neighbor classification accuracy, and majority-voting accuracy. For MIMIC-IV OOD, MIMIC-III Mortality, and MIMIC-III Readmission, we performed the same label-based analysis separately for positive and negative cases. We additionally calculated retrieval enrichment relative to the corresponding training-set class priors.

\paragraph{Layer 2: Information retention.}
For RiTeK-ADInt and RiTeK-PharmKG, we examined whether answers associated with matching source records remained explicitly present in the retrieved memory text following memory construction, compression, rewriting, or evolution. We applied case normalization, punctuation removal, whitespace normalization, and whole-word matching to compare stored memory content with reference answers. Information retention was measured conditionally on retrieving at least one answer-matching source. This analysis was not applied to TCGA-Reports or the three binary clinical prediction tasks because their compressed memory representations do not consistently preserve explicit class labels.

\paragraph{Layer 3: Evidence relevance.}
For RiTeK-ADInt and RiTeK-PharmKG, we employed Qwen3-30B-A3B-Instruct-2507 as an independent evaluator to classify test questions and retrieved source questions into three categories: same fact, related, and unrelated. The evaluator received the test question, reference answer, source question, and source answer, using deterministic decoding with temperature zero. We additionally computed cosine similarity between question embeddings generated by BAAI/bge-m3 with a maximum sequence length of 512 tokens. For TCGA-Reports, BAAI/bge-m3 was used to assess semantic similarity between test pathology reports and retrieved source reports, distinguishing matching and nonmatching cancer labels. For MIMIC-IV OOD, MIMIC-III Mortality, and MIMIC-III Readmission, we instead measured patient similarity using Jaccard similarity over normalized diagnosis-code sets, separately evaluating retrieved cases with matching and nonmatching outcome labels.

\paragraph{Memory utilization.}
Finally, we compared prediction correctness across evidence-availability and relevance categories. For biomedical question answering, we further measured the frequency with which generated answers matched labels associated with retrieved memory sources, distinguishing correct and incorrect predictions. Together, these analyses characterize the relationships among answer availability, information retention, evidence relevance, and downstream model performance.

\subsection{Construction of Controlled Conversational Memory}

To systematically evaluate the robustness of memory systems to conversation length and evidence distance, we constructed two controlled experiments from the RealMem benchmark. Each question is associated with a sequence of historical conversation sessions, categorized as relevant sessions containing information needed to answer the question or unrelated sessions that do not provide the required evidence. All transformations were performed at the session level, preserving the original content of retained sessions and the evidence required for answering each question.

\paragraph{Conversation history length.}
We constructed three conditions, Short, Medium, and Long, by selectively removing unrelated sessions preceding the earliest evidence-bearing session. Relevant sessions and all sessions occurring after the earliest evidence were preserved across conditions. In the Short condition, all preceding unrelated sessions were removed. In the Medium condition, unrelated sessions were removed sequentially from the beginning of the conversation until the remaining unrelated content was closest to half of its original token count. The Long condition retained the original conversation history without modification. This procedure produced three levels of historical context while preserving the evidence required for each question, allowing us to examine how preceding contextual information influences memory retrieval and answer generation.

\paragraph{Evidence distance.}
To isolate the effect of evidence position, we constructed Near, Medium, and Far conditions by rearranging unrelated sessions without removing or modifying their contents. All three conditions contained the same set of sessions and therefore maintained the same overall context length. In the Near condition, unrelated sessions were placed before the relevant evidence, positioning the evidence closer to the final question. In the Far condition, unrelated sessions were placed between the relevant evidence and the question, increasing the distance between them. The Medium condition distributed unrelated sessions approximately equally by token count before and after the evidence. This design isolates evidence position from context length and information availability, enabling a controlled assessment of memory sensitivity to retrieval distance.


\subsection{Hybrid Memory Retrieval and Utilization (HMRU)}

We propose Hybrid Memory Retrieval and Utilization (HMRU), a training-free framework that integrates complementary evidence retrieval and task-aware memory utilization for biomedical applications. HMRU combines dense retrieval with BM25-based lexical retrieval to capture complementary semantic and lexical evidence, followed by cross-encoder reranking to refine evidence selection. Given a query $q$, the retrieved memories are formulated as
\begin{equation}
\mathcal{M}^{*}(q)=
\operatorname{TopK}_{m\in
\mathcal{R}_{\mathrm{dense}}(q)
\cup
\mathcal{R}_{\mathrm{BM25}}(q)}
s_{\mathrm{cross}}(q,m),
\label{eq:hmru_retrieval}
\end{equation}
where $\mathcal{R}_{\mathrm{dense}}$ and $\mathcal{R}_{\mathrm{BM25}}$ denote dense and lexical retrieval, respectively, and $s_{\mathrm{cross}}(q,m)$ represents the cross-encoder relevance score.
To improve downstream evidence utilization, HMRU constructs a task-conditioned context that integrates retrieved memories with task-specific instructions:
\begin{equation}
\begin{aligned}
C_t(q) &=
[D_t;\,R(\mathcal{M}^{*}(q));\,q;\,I_t],\\
\hat{y} &= f_{\theta}(C_t(q)),
\end{aligned}
\label{eq:hmru_utilization}
\end{equation}
where $D_t$ describes the task, $I_t$ specifies evidence utilization instructions, and $R$ determines the memory representation. For biomedical knowledge reasoning, HMRU emphasizes entity--relation consistency and evidence-supported answers. For clinical prediction, it prioritizes task-relevant patient information and clinical risk factors while accounting for outcome prevalence.
We examine complementary configurations to distinguish the contributions of retrieval and utilization. HMRU-R enhances evidence selection, while HMRU-I further incorporates task-aware instructions. HMRU-C additionally adopts a compact memory representation:
\begin{equation}
R_{\mathrm{compact}}(m_i)=
[\operatorname{Input}(m_i);
\operatorname{Answer}(m_i);
\operatorname{Context}(m_i)].
\label{eq:hmru_compact}
\end{equation}
For long-term conversational memory, HMRU-T examines chronological evidence organization. All configurations operate without modifying the underlying language model parameters.


\section{Conclusion}
In this study, we systematically evaluate agentic memory systems across biomedical knowledge reasoning, clinical prediction, and long-term conversational memory tasks. We examine how memory construction, retrieval, utilization, and evolution influence downstream performance, identifying task-dependent limitations in evidence relevance, information fidelity, and long-term memory reliability. Our analyses show that retrieving relevant information does not guarantee correct utilization, semantic similarity does not necessarily correspond to clinical outcome similarity, and iterative memory updates may introduce source-unsupported content. We further demonstrate that long-term conversational memory performance cannot be explained by history length alone. To improve evidence retrieval and utilization, we introduce HMRU, a training-free framework that enhances biomedical knowledge reasoning and long-term conversational memory performance. These findings highlight the importance of evaluating memory systems beyond downstream accuracy, with particular attention to evidence accessibility, source fidelity, and effective utilization.








\bibliographystyle{unsrtnat}
\bibliography{biblatex-nature}

\appendix

\section{Supplementary Sensitivity Analyses}
\label{sec:appendix_sensitivity}

This section provides the complete results of the sensitivity analyses for biomedical memory systems. We examine changes in memory construction data volume, retrieval depth, and memory class composition while retaining the corresponding fixed test sets.

\subsection{Memory Construction Data Volume}

Table~\ref{tab:analysis_a_data_volume} reports performance across memory collections constructed from different numbers of training examples, covering biomedical knowledge reasoning, clinical prediction, and pathology classification.

\begin{table*}[!htbp]
\centering
\scriptsize
\renewcommand{\arraystretch}{1.12}
\setlength{\tabcolsep}{2.4pt}
\resizebox{0.8\textwidth}{!}{%
\begin{tabular}{@{}ll*{20}{c}@{}}
\toprule
\multirow{3}{*}{\textbf{Method}}
& \multirow{3}{*}{\textbf{Memory data}}
& \multicolumn{12}{c}{\textbf{Biomedical knowledge reasoning}}
& \multicolumn{8}{c}{\textbf{Clinical prediction and pathology classification}} \\
\cmidrule(lr){3-14}\cmidrule(l){15-22}
& & \multicolumn{6}{c}{\textbf{RiTeK-PharmKG}}
& \multicolumn{6}{c}{\textbf{RiTeK-ADint}}
& \multicolumn{2}{c}{\textbf{Mortality}}
& \multicolumn{2}{c}{\textbf{Readmission}}
& \multicolumn{2}{c}{\textbf{TCGA-Reports}}
& \multicolumn{2}{c}{\textbf{MIMIC-IV OOD}} \\
\cmidrule(lr){3-8}\cmidrule(lr){9-14}\cmidrule(lr){15-16}
\cmidrule(lr){17-18}\cmidrule(lr){19-20}\cmidrule(l){21-22}
& & $P^*$ & $R^*$ & $F_1^*$ & $P^\dagger$ & $R^\dagger$ & $F_1^\dagger$
& $P^*$ & $R^*$ & $F_1^*$ & $P^\dagger$ & $R^\dagger$ & $F_1^\dagger$
& AUPRC & AUROC & AUPRC & AUROC & AUPRC & AUROC & AUPRC & AUROC \\
\midrule
\rowcolor{gray!16}\multicolumn{22}{@{}l}{\textit{HuatuoGPT-3-27B}} \\
\multirow{3}{*}{A-MEM} & 25\% & \underline{28.14} & \underline{27.78} & \underline{27.66} & \underline{31.38} & \underline{31.00} & \underline{30.89} & \textbf{38.28} & \textbf{37.66} & \textbf{37.83} & \textbf{46.67} & \textbf{46.25} & \textbf{46.34} & 9.11 & 57.47 & \underline{58.45} & \underline{53.00} & 98.76 & 99.80 & 4.91 & 73.08 \\
 & 50\% & \textbf{33.76} & \textbf{33.48} & \textbf{33.23} & \textbf{36.52} & \textbf{36.28} & \textbf{36.05} & \underline{43.36} & \underline{43.15} & \underline{43.08} & \underline{50.23} & \underline{50.13} & \underline{50.00} & \textbf{9.67} & \underline{59.66} & \textbf{61.25} & \textbf{56.24} & \textbf{99.01} & 99.85 & 3.87 & \underline{82.96} \\
 & 100\% & \textbf{39.95} & \underline{39.23} & \textbf{39.27} & \textbf{42.18} & \textbf{41.40} & \textbf{41.51} & \textbf{53.32} & \textbf{53.08} & \textbf{52.94} & \textbf{58.84} & \textbf{58.62} & \textbf{58.53} & \textbf{9.81} & \underline{59.88} & \textbf{61.91} & \textbf{56.69} & \textbf{99.06} & \textbf{99.93} & 4.94 & 82.16 \\
\addlinespace[2pt]
\multirow{3}{*}{MemZero} & 25\% & 26.25 & 25.43 & 25.64 & 29.22 & 28.39 & 28.60 & 30.19 & 29.88 & 29.94 & 38.98 & 38.78 & 38.78 & \underline{9.33} & 53.16 & 55.62 & 51.64 & \underline{98.99} & 99.85 & 3.01 & 72.07 \\
 & 50\% & 28.53 & 28.35 & 28.13 & 30.90 & 30.74 & 30.50 & 34.54 & 34.30 & 34.27 & 42.39 & 42.28 & 42.18 & 7.95 & 52.54 & 57.94 & 51.74 & 98.58 & \underline{99.88} & \underline{5.29} & 76.53 \\
 & 100\% & 31.22 & 31.07 & 30.84 & 33.96 & 33.76 & 33.57 & 43.57 & 43.43 & 43.31 & 50.05 & 49.99 & 49.84 & 8.29 & 54.21 & 57.98 & 52.71 & 98.64 & 99.87 & 6.95 & 78.94 \\
\addlinespace[2pt]
\multirow{3}{*}{FullContext} & 25\% & 26.98 & 25.76 & 26.10 & 29.72 & 28.63 & 28.94 & 36.51 & \underline{35.96} & 36.11 & 44.25 & 43.96 & 43.99 & \textbf{10.85} & \textbf{60.21} & 55.13 & 50.60 & 98.50 & \underline{99.86} & \textbf{8.83} & \textbf{85.28} \\
 & 50\% & 26.84 & 26.30 & 26.20 & 30.99 & 30.48 & 30.42 & 32.05 & 31.71 & 31.71 & 38.86 & 38.65 & 38.63 & 9.03 & 57.59 & 58.03 & 52.34 & \underline{98.83} & \textbf{99.89} & \textbf{7.26} & \textbf{83.64} \\
 & 100\% & 27.55 & 26.65 & 26.80 & 30.25 & 29.51 & 29.64 & 37.34 & 36.58 & 36.83 & 45.30 & 44.68 & 44.90 & \underline{9.68} & 59.19 & 57.93 & 52.55 & 98.88 & \underline{99.88} & \underline{8.23} & \underline{82.76} \\
\addlinespace[2pt]
\multirow{3}{*}{NaiveRAG} & 25\% & \textbf{28.71} & \textbf{28.35} & \textbf{28.13} & \textbf{31.77} & \textbf{31.40} & \textbf{31.24} & \underline{38.07} & \textbf{37.66} & \underline{37.76} & \underline{46.63} & \underline{46.21} & \underline{46.32} & 9.30 & \underline{58.46} & \textbf{59.55} & \textbf{53.20} & \textbf{99.10} & \textbf{99.88} & \underline{7.70} & 82.80 \\
 & 50\% & \underline{32.32} & \underline{32.42} & \underline{31.93} & \underline{35.10} & \underline{35.29} & \underline{34.76} & \textbf{44.50} & \textbf{44.05} & \textbf{44.09} & \textbf{51.04} & \textbf{50.78} & \textbf{50.74} & \underline{9.44} & \textbf{60.49} & \underline{59.28} & \underline{54.98} & 98.75 & 99.82 & 5.22 & 81.65 \\
 & 100\% & \underline{39.42} & \textbf{39.33} & \underline{39.04} & \underline{41.48} & \underline{41.36} & \underline{41.07} & \underline{52.07} & \underline{51.63} & \underline{51.63} & \underline{57.69} & \underline{57.34} & \underline{57.36} & \textbf{9.81} & \textbf{61.98} & \underline{60.76} & \underline{55.79} & \underline{98.97} & \textbf{99.93} & \textbf{15.61} & \textbf{85.47} \\
\addlinespace[2pt]
\multirow{3}{*}{LightMem} & 25\% & 27.75 & 27.06 & 27.13 & 30.82 & 30.15 & 30.24 & 33.61 & 33.40 & 33.40 & 41.12 & 40.99 & 40.95 & 8.96 & 54.25 & 54.88 & 49.01 & 98.74 & 99.85 & 5.21 & \underline{83.54} \\
 & 50\% & 31.48 & 31.31 & 31.05 & 34.05 & 33.96 & 33.68 & 38.80 & 38.66 & 38.55 & 45.58 & 45.45 & 45.37 & 8.79 & 55.38 & 56.87 & 51.34 & 98.70 & \textbf{99.89} & 4.28 & 78.41 \\
 & 100\% & 37.05 & 36.84 & 36.62 & 39.25 & 38.89 & 38.78 & 50.21 & 49.76 & 49.76 & 55.06 & 54.88 & 54.77 & 9.62 & 57.80 & 59.17 & 53.91 & 98.47 & 99.81 & 3.42 & 79.11 \\
\midrule
\rowcolor{gray!16}\multicolumn{22}{@{}l}{\textit{Qwen3-30B-A3B-Instruct-2507}} \\
\multirow{3}{*}{A-MEM} & 25\% & 22.31 & \underline{25.64} & 23.17 & 25.20 & \underline{29.25} & \underline{26.36} & 28.27 & \underline{34.34} & 30.11 & 35.05 & \underline{44.02} & 37.84 & 9.59 & \underline{54.25} & \textbf{56.72} & 48.58 & \underline{97.77} & \underline{99.77} & 3.74 & 73.95 \\
 & 50\% & \underline{27.43} & \underline{32.10} & \underline{28.72} & \underline{29.83} & \underline{34.98} & \underline{31.29} & \textbf{36.45} & \textbf{43.02} & \textbf{38.41} & \underline{42.22} & \textbf{51.13} & \underline{44.92} & \underline{9.20} & \underline{53.36} & \underline{56.98} & \underline{51.70} & 97.68 & \underline{99.85} & \underline{4.06} & 75.72 \\
 & 100\% & \underline{31.79} & \underline{36.23} & \underline{33.00} & \underline{33.95} & \underline{38.64} & \underline{35.28} & \underline{44.35} & \textbf{50.80} & \underline{46.20} & \underline{50.69} & \textbf{58.69} & \underline{53.18} & \underline{9.37} & 51.17 & 55.01 & 49.76 & \textbf{98.39} & \underline{99.81} & \textbf{4.13} & \textbf{81.66} \\
\addlinespace[2pt]
\multirow{3}{*}{MemZero} & 25\% & 17.76 & 22.50 & 19.11 & 20.40 & 25.89 & 22.00 & 19.37 & 24.86 & 21.04 & 26.21 & 35.70 & 29.12 & 8.06 & 50.34 & \underline{56.50} & \underline{50.52} & 93.74 & 99.44 & 3.34 & 75.42 \\
 & 50\% & 19.06 & 24.14 & 20.45 & 20.94 & 26.64 & 22.62 & 25.11 & 32.12 & 27.14 & 30.60 & 40.22 & 33.54 & 8.11 & 50.64 & 56.08 & 50.76 & 96.53 & 99.65 & 2.16 & 72.24 \\
 & 100\% & 23.76 & 29.20 & 25.29 & 25.26 & 31.31 & 27.01 & 28.75 & 35.79 & 30.80 & 34.82 & 44.95 & 37.86 & 7.54 & 51.35 & \underline{57.13} & \underline{52.33} & 96.76 & 99.75 & 2.67 & 74.60 \\
\addlinespace[2pt]
\multirow{3}{*}{FullContext} & 25\% & \underline{23.13} & 25.29 & \underline{23.47} & \underline{25.32} & 28.08 & 25.93 & \underline{29.82} & 33.02 & \underline{30.68} & \underline{37.34} & 42.76 & \underline{38.95} & \textbf{10.00} & \textbf{54.83} & 56.13 & 49.30 & 95.08 & 99.59 & \textbf{7.23} & \underline{76.58} \\
 & 50\% & 21.48 & 23.67 & 21.90 & 24.82 & 27.49 & 25.40 & 25.17 & 30.57 & 26.74 & 32.34 & 39.83 & 34.64 & 8.42 & 53.08 & 56.71 & 50.65 & \underline{97.70} & \textbf{99.87} & 3.43 & \underline{79.02} \\
 & 100\% & 24.11 & 26.16 & 24.42 & 26.53 & 29.12 & 27.07 & 29.96 & 34.09 & 31.04 & 37.46 & 44.43 & 39.57 & \textbf{10.27} & \underline{52.90} & 55.50 & 49.78 & 95.41 & 99.43 & 3.52 & 73.96 \\
\addlinespace[2pt]
\multirow{3}{*}{NaiveRAG} & 25\% & \textbf{24.15} & \textbf{27.57} & \textbf{25.01} & \textbf{26.93} & \textbf{31.08} & \textbf{28.07} & \textbf{31.92} & \textbf{36.17} & \textbf{33.12} & \textbf{39.94} & \textbf{46.88} & \textbf{42.04} & \underline{9.79} & 54.16 & 54.82 & \textbf{50.72} & \textbf{97.85} & 99.65 & \underline{3.88} & \textbf{77.82} \\
 & 50\% & \textbf{28.88} & \textbf{33.72} & \textbf{30.16} & \textbf{31.32} & \textbf{36.54} & \textbf{32.76} & \underline{36.26} & \underline{42.15} & \underline{37.98} & \textbf{43.27} & \underline{50.98} & \textbf{45.66} & \textbf{9.71} & \textbf{61.01} & \textbf{57.68} & \textbf{52.45} & \textbf{98.29} & 99.75 & 2.61 & 74.94 \\
 & 100\% & \textbf{35.24} & \textbf{40.29} & \textbf{36.53} & \textbf{36.87} & \textbf{41.87} & \textbf{38.18} & \textbf{46.14} & \underline{50.17} & \textbf{47.29} & \textbf{52.68} & \underline{57.93} & \textbf{54.26} & 9.02 & \textbf{56.57} & \textbf{57.89} & \textbf{54.70} & \underline{98.36} & \textbf{99.90} & \underline{3.92} & \underline{78.59} \\
\addlinespace[2pt]
\multirow{3}{*}{LightMem} & 25\% & 21.91 & 25.59 & 22.91 & 24.24 & 28.53 & 25.46 & 27.58 & 32.88 & 29.21 & 34.52 & 42.75 & 37.16 & 8.98 & 53.85 & 55.08 & 49.54 & 97.10 & \textbf{99.81} & 2.50 & 72.03 \\
 & 50\% & 25.40 & 29.52 & 26.43 & 27.65 & 32.29 & 28.88 & 32.64 & 38.66 & 34.39 & 38.10 & 46.43 & 40.66 & 7.80 & 50.44 & 54.91 & 50.12 & 97.27 & 99.72 & \textbf{6.63} & \textbf{80.54} \\
 & 100\% & 29.67 & 33.77 & 30.73 & 31.31 & 35.57 & 32.47 & 39.64 & 45.30 & 41.16 & 45.29 & 52.65 & 47.49 & 8.88 & 52.19 & 54.76 & 49.76 & 97.76 & \underline{99.81} & 3.25 & 77.61 \\
\bottomrule
\end{tabular}%
}
\caption{\textbf{Effect of memory construction data volume across six biomedical tasks.}
Nested memory-construction subsets contain 250, 500, and 1,000 training samples,
corresponding to 25\%, 50\%, and 100\% of the full subset. The test sets and other
experimental settings are fixed. RiTeK tasks report precision ($P$), recall ($R$),
and F1 ($F_1$) using exact match ($^*$) and ROUGE-1 ($^\dagger$).
Mortality and readmission use MIMIC-III. TCGA-Reports results are macro-averaged.
Clinical prediction and pathology classification report AUPRC and AUROC.
All results are percentages. Within each model and memory data volume,
\textbf{bold} and \underline{underlined} values indicate the best and second-best
scores among memory approaches, respectively.}
\label{tab:analysis_a_data_volume}
\end{table*}
\FloatBarrier

\subsection{Retrieval Depth}

Table~\ref{tab:analysis_b_topk} presents the results obtained with different numbers of retrieved memory items for each model and memory method.

\begin{table*}[!htbp]
\centering
\scriptsize
\renewcommand{\arraystretch}{1.12}
\setlength{\tabcolsep}{2.4pt}
\resizebox{\textwidth}{!}{%
\begin{tabular}{@{}ll*{20}{c}@{}}
\toprule
\multirow{3}{*}{\textbf{Method}}
& \multirow{3}{*}{\textbf{Top-$k$}}
& \multicolumn{12}{c}{\textbf{Biomedical knowledge reasoning}}
& \multicolumn{8}{c}{\textbf{Clinical prediction and pathology classification}} \\
\cmidrule(lr){3-14}\cmidrule(l){15-22}
& & \multicolumn{6}{c}{\textbf{RiTeK-PharmKG}}
& \multicolumn{6}{c}{\textbf{RiTeK-ADint}}
& \multicolumn{2}{c}{\textbf{Mortality}}
& \multicolumn{2}{c}{\textbf{Readmission}}
& \multicolumn{2}{c}{\textbf{TCGA-Reports}}
& \multicolumn{2}{c}{\textbf{MIMIC-IV OOD}} \\
\cmidrule(lr){3-8}\cmidrule(lr){9-14}\cmidrule(lr){15-16}
\cmidrule(lr){17-18}\cmidrule(lr){19-20}\cmidrule(l){21-22}
& & $P^*$ & $R^*$ & $F_1^*$ & $P^\dagger$ & $R^\dagger$ & $F_1^\dagger$
& $P^*$ & $R^*$ & $F_1^*$ & $P^\dagger$ & $R^\dagger$ & $F_1^\dagger$
& AUPRC & AUROC & AUPRC & AUROC & AUPRC & AUROC & AUPRC & AUROC \\
\midrule
\rowcolor{gray!16}\multicolumn{22}{@{}l}{\textit{Qwen3-30B-A3B-Instruct-2507}} \\
\multirow{4}{*}{A-MEM} & 1 & \underline{26.03} & \textbf{31.02} & \underline{27.33} & \underline{28.72} & \underline{34.46} & \underline{30.28} & \underline{34.98} & \textbf{41.94} & \underline{36.71} & \underline{40.28} & \textbf{50.02} & \underline{42.94} & \underline{8.99} & \textbf{56.72} & \underline{56.17} & \underline{52.02} & \underline{96.02} & \underline{99.56} & \textbf{6.42} & \textbf{78.18} \\
 & 3 & \underline{29.14} & \underline{33.46} & \underline{30.32} & \underline{31.29} & \underline{36.14} & \underline{32.68} & \underline{41.13} & \underline{46.23} & \underline{42.57} & \underline{46.92} & \underline{54.37} & \underline{49.09} & \underline{8.43} & \underline{54.76} & 55.54 & 50.89 & \underline{97.03} & \underline{99.63} & \textbf{4.10} & 73.39 \\
 & 5 & \underline{30.45} & \underline{35.57} & \underline{31.81} & \underline{32.63} & \underline{38.03} & \underline{34.12} & \underline{43.25} & \underline{49.20} & \underline{44.95} & \underline{49.01} & \underline{57.27} & \underline{51.50} & \underline{8.86} & \underline{54.00} & \underline{56.98} & \underline{52.28} & \underline{97.97} & \underline{99.77} & 3.08 & \underline{74.98} \\
 & 10 & \underline{31.79} & \underline{36.23} & \underline{33.00} & \underline{33.95} & \underline{38.64} & \underline{35.28} & \underline{44.35} & \textbf{50.80} & \underline{46.20} & \underline{50.69} & \textbf{58.69} & \underline{53.18} & \textbf{9.37} & 51.17 & 55.01 & 49.76 & \textbf{98.39} & \underline{99.81} & \textbf{4.13} & \textbf{81.66} \\
\addlinespace[2pt]
\multirow{4}{*}{MemZero} & 1 & 18.99 & 22.97 & 20.13 & 21.81 & 26.43 & 23.16 & 27.40 & 32.54 & 28.82 & 32.35 & 39.98 & 34.53 & 8.85 & 53.43 & \textbf{57.33} & \textbf{53.15} & 94.06 & 99.39 & \underline{6.27} & 74.18 \\
 & 3 & 21.42 & 26.44 & 22.82 & 23.40 & 29.02 & 25.02 & 28.01 & 34.41 & 29.87 & 34.02 & 43.65 & 36.94 & 7.75 & 51.29 & \textbf{57.37} & \underline{52.17} & 95.49 & 99.59 & 2.22 & 72.77 \\
 & 5 & 22.33 & 27.31 & 23.78 & 24.21 & 29.88 & 25.91 & 28.55 & 34.20 & 30.21 & 34.19 & 42.78 & 36.80 & 8.03 & 49.78 & 55.83 & 51.03 & 95.62 & 99.60 & 2.12 & 72.61 \\
 & 10 & 23.76 & 29.20 & 25.29 & 25.26 & 31.31 & 27.01 & 28.75 & 35.79 & 30.80 & 34.82 & 44.95 & 37.86 & 7.54 & 51.35 & \underline{57.13} & \underline{52.33} & 96.76 & 99.75 & 2.67 & 74.60 \\
\addlinespace[2pt]
\multirow{4}{*}{LightMem} & 1 & 23.09 & 27.45 & 24.25 & 25.97 & 31.09 & 27.35 & 31.66 & 36.27 & 32.83 & 36.81 & 44.25 & 38.92 & 7.28 & 49.87 & 54.97 & 49.35 & 94.74 & 99.28 & 3.56 & 74.16 \\
 & 3 & 26.81 & 31.02 & 28.08 & 28.88 & 33.56 & 30.30 & 35.50 & 41.56 & 37.19 & 40.99 & 49.37 & 43.54 & 7.86 & 50.37 & 54.63 & 49.39 & 96.70 & 99.37 & \underline{3.41} & \textbf{77.47} \\
 & 5 & 28.54 & 33.02 & 29.82 & 30.43 & 35.16 & 31.82 & 37.54 & 42.60 & 38.91 & 43.62 & 51.02 & 45.83 & 8.66 & 53.20 & 54.54 & 49.57 & 97.10 & 99.59 & \textbf{4.46} & 74.65 \\
 & 10 & 29.67 & 33.77 & 30.73 & 31.31 & 35.57 & 32.47 & 39.64 & 45.30 & 41.16 & 45.29 & 52.65 & 47.49 & 8.88 & \underline{52.19} & 54.76 & 49.76 & 97.76 & \underline{99.81} & 3.25 & 77.61 \\
\addlinespace[2pt]
\multirow{4}{*}{NaiveRAG} & 1 & \textbf{27.30} & \underline{30.97} & \textbf{28.23} & \textbf{30.47} & \textbf{34.91} & \textbf{31.65} & \textbf{36.02} & \underline{41.04} & \textbf{37.34} & \textbf{41.67} & \underline{49.10} & \textbf{43.80} & \textbf{9.33} & \underline{55.87} & 55.76 & 51.55 & \textbf{96.31} & \textbf{99.65} & 3.68 & \underline{76.46} \\
 & 3 & \textbf{31.19} & \textbf{35.73} & \textbf{32.36} & \textbf{33.04} & \textbf{37.87} & \textbf{34.36} & \textbf{42.48} & \textbf{47.89} & \textbf{43.98} & \textbf{48.73} & \textbf{55.75} & \textbf{50.82} & \textbf{10.16} & \textbf{57.59} & \underline{56.79} & \textbf{53.00} & \textbf{98.33} & \textbf{99.85} & 2.55 & \underline{73.53} \\
 & 5 & \textbf{33.31} & \textbf{37.90} & \textbf{34.47} & \textbf{34.86} & \textbf{39.68} & \textbf{36.12} & \textbf{45.07} & \textbf{49.55} & \textbf{46.24} & \textbf{51.58} & \textbf{58.04} & \textbf{53.52} & \textbf{9.21} & \textbf{57.90} & \textbf{58.08} & \textbf{53.61} & \textbf{98.13} & \textbf{99.85} & \underline{3.23} & \textbf{76.71} \\
 & 10 & \textbf{35.24} & \textbf{40.29} & \textbf{36.53} & \textbf{36.87} & \textbf{41.87} & \textbf{38.18} & \textbf{46.14} & \underline{50.17} & \textbf{47.29} & \textbf{52.68} & \underline{57.93} & \textbf{54.26} & \underline{9.02} & \textbf{56.57} & \textbf{57.89} & \textbf{54.70} & \underline{98.36} & \textbf{99.90} & \underline{3.92} & \underline{78.59} \\
\midrule
\rowcolor{gray!16}\multicolumn{22}{@{}l}{\textit{HuatuoGPT-3-27B}} \\
\multirow{4}{*}{A-MEM} & 1 & \underline{29.49} & \underline{28.82} & \underline{28.93} & \underline{32.69} & \underline{32.07} & \underline{32.16} & \textbf{43.27} & \textbf{42.95} & \textbf{42.84} & \textbf{48.64} & \textbf{48.45} & \textbf{48.32} & \textbf{9.97} & \textbf{58.47} & 59.43 & \underline{55.94} & 98.86 & 99.81 & \underline{4.67} & \textbf{77.97} \\
 & 3 & \textbf{35.83} & \textbf{35.42} & \textbf{35.34} & \textbf{38.33} & \textbf{37.88} & \textbf{37.84} & \textbf{50.93} & \textbf{50.48} & \textbf{50.45} & \textbf{56.20} & \textbf{55.90} & \textbf{55.87} & 9.70 & \underline{60.07} & 59.61 & \underline{55.41} & 98.66 & 99.83 & 3.57 & \underline{79.35} \\
 & 5 & \textbf{37.65} & \textbf{37.23} & \textbf{37.15} & \textbf{40.18} & \textbf{39.76} & \textbf{39.70} & \textbf{53.01} & \textbf{52.56} & \textbf{52.52} & \textbf{58.66} & \textbf{58.36} & \textbf{58.33} & \textbf{9.49} & \underline{58.10} & 59.63 & 55.74 & \underline{98.94} & \textbf{99.93} & 5.31 & \underline{83.75} \\
 & 10 & \textbf{39.95} & \underline{39.23} & \textbf{39.27} & \textbf{42.18} & \textbf{41.40} & \textbf{41.51} & \textbf{53.32} & \textbf{53.08} & \textbf{52.94} & \textbf{58.84} & \textbf{58.62} & \textbf{58.53} & \textbf{9.81} & \underline{59.88} & \textbf{61.91} & \textbf{56.69} & \textbf{99.06} & \textbf{99.93} & 4.94 & \underline{82.16} \\
\addlinespace[2pt]
\multirow{4}{*}{MemZero} & 1 & 23.64 & 23.20 & 23.22 & 27.59 & 27.19 & 27.17 & 35.37 & 35.06 & 35.06 & 41.49 & 41.52 & 41.29 & \underline{9.82} & 56.25 & 58.86 & 53.95 & \underline{98.93} & \underline{99.89} & 3.92 & 70.89 \\
 & 3 & 27.24 & 26.87 & 26.78 & 30.50 & 30.14 & 30.06 & 39.21 & 39.00 & 38.93 & 46.18 & 46.11 & 46.00 & \underline{9.71} & 57.70 & 57.93 & 50.96 & 98.54 & \underline{99.89} & \textbf{6.34} & 78.81 \\
 & 5 & 29.20 & 29.21 & 28.89 & 32.01 & 31.97 & 31.68 & 42.84 & 42.74 & 42.60 & 49.62 & 49.69 & 49.44 & 9.12 & 53.67 & 57.36 & 51.24 & 98.50 & \underline{99.87} & \textbf{7.43} & 79.50 \\
 & 10 & 31.22 & 31.07 & 30.84 & 33.96 & 33.76 & 33.57 & 43.57 & 43.43 & 43.31 & 50.05 & 49.99 & 49.84 & 8.29 & 54.21 & 57.98 & 52.71 & 98.64 & \underline{99.87} & \underline{6.95} & 78.94 \\
\addlinespace[2pt]
\multirow{4}{*}{LightMem} & 1 & 29.14 & 28.48 & 28.57 & 32.12 & 31.55 & 31.59 & \underline{41.08} & \underline{40.46} & \underline{40.66} & \underline{46.54} & \underline{46.16} & \underline{46.25} & 8.78 & \underline{57.70} & \textbf{60.23} & 55.20 & \textbf{99.17} & \textbf{99.92} & \textbf{4.90} & 71.69 \\
 & 3 & 33.04 & 32.73 & 32.67 & 35.76 & 35.29 & 35.33 & 47.61 & 46.96 & 47.06 & 52.38 & 52.06 & 52.02 & 7.99 & 54.47 & \underline{60.97} & 55.25 & \underline{98.87} & 99.88 & 4.02 & 78.57 \\
 & 5 & 34.41 & 33.85 & 33.89 & 36.48 & 35.89 & 35.96 & 47.93 & 47.58 & 47.54 & 53.82 & 53.52 & 53.52 & 7.53 & 52.99 & \textbf{61.32} & \underline{56.09} & 98.73 & \underline{99.87} & 3.28 & 76.99 \\
 & 10 & 37.05 & 36.84 & 36.62 & 39.25 & 38.89 & 38.78 & 50.21 & 49.76 & 49.76 & 55.06 & 54.88 & 54.77 & \underline{9.62} & 57.80 & 59.17 & 53.91 & 98.47 & 99.81 & 3.42 & 79.11 \\
\addlinespace[2pt]
\multirow{4}{*}{NaiveRAG} & 1 & \textbf{29.80} & \textbf{29.57} & \textbf{29.52} & \textbf{33.56} & \textbf{33.16} & \textbf{33.17} & 40.56 & 40.15 & 40.18 & 45.57 & 45.35 & 45.30 & 8.63 & 56.55 & \underline{59.75} & \textbf{56.10} & 98.64 & 99.87 & 3.99 & \underline{74.58} \\
 & 3 & \underline{33.92} & \underline{33.69} & \underline{33.52} & \underline{36.60} & \underline{36.29} & \underline{36.18} & \underline{48.76} & \underline{48.31} & \underline{48.31} & \underline{54.39} & \underline{54.16} & \underline{54.12} & \textbf{10.41} & \textbf{62.99} & \textbf{61.06} & \textbf{55.72} & \textbf{98.90} & \textbf{99.92} & \underline{5.26} & \textbf{87.62} \\
 & 5 & \underline{36.58} & \underline{36.59} & \underline{36.22} & \underline{38.86} & \underline{38.87} & \underline{38.52} & \underline{50.00} & \underline{49.65} & \underline{49.62} & \underline{55.46} & \underline{55.28} & \underline{55.20} & \underline{9.25} & \textbf{60.59} & \underline{60.94} & \textbf{56.71} & \textbf{99.03} & \textbf{99.93} & \underline{5.84} & \textbf{88.67} \\
 & 10 & \underline{39.42} & \textbf{39.33} & \underline{39.04} & \underline{41.48} & \underline{41.36} & \underline{41.07} & \underline{52.07} & \underline{51.63} & \underline{51.63} & \underline{57.69} & \underline{57.34} & \underline{57.36} & \textbf{9.81} & \textbf{61.98} & \underline{60.76} & \underline{55.79} & \underline{98.97} & \textbf{99.93} & \textbf{15.61} & \textbf{85.47} \\
\bottomrule
\end{tabular}%
}
\caption{\textbf{Effect of retrieval depth on biomedical memory performance.}
We evaluate four memory methods (A-MEM, MemZero, LightMem, and NaiveRAG)
with two language-model backbones across six biomedical tasks. For each
benchmark, the same 1,000 samples are used to construct the memory or
retrieval corpus, and the number of retrieved memory items varies over
$k\in\{1,3,5,10\}$. The test sets and other experimental settings are held
fixed. RiTeK tasks report precision ($P$), recall ($R$), and F1 ($F_1$)
using exact match ($^*$) and ROUGE-1 ($^\dagger$). Mortality and readmission
are evaluated on MIMIC-III; TCGA-Reports results are macro-averaged.
Clinical prediction and pathology classification report AUPRC and AUROC.
All results are percentages. Within each model and retrieval depth,
\textbf{bold} and \underline{underlined} values indicate the best and
second-best scores among the four memory approaches, respectively. The
Top-10 results reproduce the corresponding full-data baseline results.}
\label{tab:analysis_b_topk}
\end{table*}
\FloatBarrier

\subsection{Memory Class Distribution}

Table~\ref{tab:analysis_c_class_distribution} reports the effects of varying positive-to-negative class composition in the memory collection for mortality and readmission prediction.

\begin{table*}[!htbp]
\centering
\scriptsize
\renewcommand{\arraystretch}{1.12}
\setlength{\tabcolsep}{5.5pt}
\resizebox{0.65\textwidth}{!}{%
\begin{tabular}{llcccc}
\toprule
\multirow{2}{*}{\textbf{Method}}
& \multirow{2}{*}{\textbf{Pos:Neg}}
& \multicolumn{2}{c}{\textbf{Mortality--MIMIC-III}}
& \multicolumn{2}{c}{\textbf{Readmission--MIMIC-III}} \\
\cmidrule(lr){3-4}\cmidrule(lr){5-6}
& & \textbf{AUPRC} & \textbf{AUROC}
& \textbf{AUPRC} & \textbf{AUROC} \\
\midrule

\rowcolor{gray!15}
\multicolumn{6}{l}{\textit{Qwen3-30B-A3B-Instruct-2507}} \\
\addlinespace[1pt]

\multirow{4}{*}{A-MEM}
& 10:90 & \underline{8.57} & \textbf{54.49} & \underline{55.06} & \underline{49.29} \\
& 25:75 & \textbf{9.67} & \underline{54.66} & \textbf{57.92} & \textbf{52.23} \\
& 50:50 & 7.11 & 49.38 & \underline{56.82} & \underline{51.30} \\
& Natural & \underline{9.37} & 51.17 & 55.01 & 49.76 \\
\addlinespace[3pt]

\multirow{4}{*}{MemZero}
& 10:90 & 7.37 & 49.38 & 54.51 & 48.59 \\
& 25:75 & 7.16 & 50.46 & 53.65 & 47.97 \\
& 50:50 & 7.46 & 50.73 & \textbf{56.88} & \textbf{52.46} \\
& Natural & 7.54 & 51.35 & \underline{57.13} & \underline{52.33} \\
\addlinespace[3pt]

\multirow{4}{*}{FullContext}
& 10:90 & 7.68 & 51.34 & 54.84 & 48.37 \\
& 25:75 & 7.72 & 50.02 & 55.15 & 48.52 \\
& 50:50 & \textbf{11.17} & \textbf{56.01} & 54.37 & 47.02 \\
& Natural & \textbf{10.27} & \underline{52.90} & 55.50 & 49.78 \\
\addlinespace[3pt]

\multirow{4}{*}{NaiveRAG}
& 10:90 & 7.77 & 50.69 & 53.72 & 47.86 \\
& 25:75 & 8.44 & \textbf{55.28} & 55.29 & 50.09 \\
& 50:50 & 7.93 & \underline{52.45} & 56.44 & 49.67 \\
& Natural & 9.02 & \textbf{56.57} & \textbf{57.89} & \textbf{54.70} \\
\addlinespace[3pt]

\multirow{4}{*}{LightMem}
& 10:90 & \textbf{10.77} & \underline{52.72} & \textbf{56.14} & \textbf{50.75} \\
& 25:75 & \underline{9.16} & 53.00 & \underline{55.65} & \underline{50.88} \\
& 50:50 & \underline{8.13} & 51.98 & 54.71 & 50.33 \\
& Natural & 8.88 & 52.19 & 54.76 & 49.76 \\

\midrule

\rowcolor{gray!15}
\multicolumn{6}{l}{\textit{HuatuoGPT-3-27B}} \\
\addlinespace[1pt]

\multirow{4}{*}{A-MEM}
& 10:90 & \textbf{10.17} & \textbf{60.79} & 56.09 & 50.02 \\
& 25:75 & \textbf{10.25} & 56.24 & \underline{58.42} & \underline{54.48} \\
& 50:50 & 10.07 & \underline{58.34} & \textbf{60.37} & \textbf{57.16} \\
& Natural & \textbf{9.81} & \underline{59.88} & \textbf{61.91} & \textbf{56.69} \\
\addlinespace[3pt]

\multirow{4}{*}{MemZero}
& 10:90 & 8.40 & 52.58 & \textbf{58.88} & \textbf{54.62} \\
& 25:75 & 9.25 & \underline{57.77} & 56.12 & 53.13 \\
& 50:50 & 8.99 & 56.67 & 53.26 & 48.78 \\
& Natural & 8.29 & 54.21 & 57.98 & 52.71 \\
\addlinespace[3pt]

\multirow{4}{*}{FullContext}
& 10:90 & \underline{9.88} & 56.05 & 56.05 & 50.85 \\
& 25:75 & 9.35 & \underline{57.77} & 57.06 & 51.50 \\
& 50:50 & \textbf{10.58} & \textbf{61.17} & 55.35 & 50.40 \\
& Natural & \underline{9.68} & 59.19 & 57.93 & 52.55 \\
\addlinespace[3pt]

\multirow{4}{*}{NaiveRAG}
& 10:90 & 8.48 & 57.53 & 56.01 & 50.33 \\
& 25:75 & \underline{9.87} & \textbf{59.04} & \textbf{58.75} & \textbf{54.96} \\
& 50:50 & 9.28 & 57.96 & 55.93 & 51.90 \\
& Natural & \textbf{9.81} & \textbf{61.98} & \underline{60.76} & \underline{55.79} \\
\addlinespace[3pt]

\multirow{4}{*}{LightMem}
& 10:90 & 9.72 & \underline{58.80} & \underline{58.32} & \underline{52.34} \\
& 25:75 & 9.78 & 54.90 & 54.68 & 47.82 \\
& 50:50 & \underline{10.20} & 57.32 & \underline{58.58} & \underline{52.89} \\
& Natural & 9.62 & 57.80 & 59.17 & 53.91 \\

\bottomrule
\end{tabular}%
}

\caption{
\textbf{Effect of class distribution on biomedical memory performance.}
Five memory methods are evaluated using two language-model backbones on MIMIC-III mortality and readmission prediction. For each task, 1,000 samples are selected for memory construction or context preparation under positive-to-negative ratios of 10:90, 25:75, and 50:50, or under the natural distribution without explicit class balancing. Test sets retain their original class distributions, and other experimental settings are held fixed. Performance is reported as AUPRC and AUROC in percentages. Within each backbone and class-ratio setting, the best and second-best results across methods are shown in bold and underlined, respectively. Natural rows reproduce the corresponding baseline results.
}

\label{tab:analysis_c_class_distribution}
\end{table*}
\FloatBarrier

\section{Supplementary Prompt Templates}
\label{sec:appendix_prompts}

This section provides the prompt templates used to examine task-aware evidence utilization and memory representation. The biomedical prompts compare the original and compact memory formats, while the RealMem prompts contrast relevance-ordered memory utilization with chronological state tracking.

\subsection{Task-Aware Biomedical Memory Utilization}

Figures~\ref{fig:tamua_instruction} and~\ref{fig:tamua_compact} show the HMRU-I and HMRU-C prompt templates, respectively. The latter retains task-specific instructions while simplifying the presentation of retrieved memories.

\begin{figure*}[p]
\centering

\begin{biomemprompt}{
    HMRU-I: Task-Aware Memory Utilization
}

You are an expert biomedical assistant. Your task is to
answer the current question using relevant training memories
and task-specific evidence utilization rules.

\medskip
\textbf{Task Description}

\promptvar{task description}

\medskip
\textbf{Retrieved Training Memory}

The following memories are ranked from most to least
relevant to the current query.

\begin{tcolorbox}[
    colback=gray!8,
    colframe=gray!35,
    arc=5pt,
    boxrule=0.4pt
]
\textbf{Memory Content:}
\promptvar{input and answer}

\textbf{Memory Context:}
\promptvar{generated context}

\textbf{Memory Keywords:}
\promptvar{keywords}

\textbf{Memory Tags:}
\promptvar{tags}

\textbf{Talk Start Time:}
\promptvar{timestamp}

\medskip
\textit{Additional retrieved memories follow.}
\end{tcolorbox}

\textbf{Current Question or Patient}

\promptvar{query or patient diagnosis history}

\medskip
\TAMUARules

\medskip
\textbf{Output Format}

\begin{tcolorbox}[
    colback=gray!8,
    colframe=gray!35,
    arc=5pt,
    boxrule=0.4pt
]
\texttt{Biomedical QA: }
\promptvar{entity name}

\texttt{Clinical prediction: true / false}

\medskip
\textbf{Answer:}
\end{tcolorbox}

\end{biomemprompt}

\caption{
\textbf{Task-aware memory utilization instructions in HMRU-I.}
The adapter applies task-specific evidence utilization rules
while preserving the original memory representation.
}
\label{fig:tamua_instruction}
\end{figure*}
\FloatBarrier

\begin{figure*}[p]
\centering

\begin{biomemprompt}{
    HMRU-C: Compact Task-Aware Memory Utilization
}

You are an expert biomedical assistant. Your task is to
answer the current question using relevant training memories
and task-specific evidence utilization rules.

\medskip
\textbf{Task Description}

\promptvar{task description}

\medskip
\textbf{Retrieved Training Memory}

The following memories are ranked from most to least
relevant to the current query. Each memory retains its
input, answer, and contextual summary.

\begin{tcolorbox}[
    colback=gray!8,
    colframe=gray!35,
    arc=5pt,
    boxrule=0.4pt
]
\textbf{[M1]}

\textbf{Input:} \promptvar{training input 1}

\textbf{Answer:} \promptvar{training answer 1}

\textbf{Summary:} \promptvar{context 1}

\medskip
\textbf{[M2]}

\textbf{Input:} \promptvar{training input 2}

\textbf{Answer:} \promptvar{training answer 2}

\textbf{Summary:} \promptvar{context 2}

\medskip
\textit{Additional retrieved memories follow.}
\end{tcolorbox}

\textbf{Current Question or Patient}

\promptvar{query or patient diagnosis history}

\medskip
\TAMUARules

\medskip
\textbf{Output Format}

\begin{tcolorbox}[
    colback=gray!8,
    colframe=gray!35,
    arc=5pt,
    boxrule=0.4pt
]
\texttt{Biomedical QA: }
\promptvar{entity name}

\texttt{Clinical prediction: true / false}

\medskip
\textbf{Answer:}
\end{tcolorbox}

\end{biomemprompt}

\caption{
\textbf{Compact memory representation in HMRU-C.}
The adapter retains the task-aware utilization instructions
of HMRU-I while removing auxiliary memory metadata,
including keywords, tags, and timestamps.
}
\label{fig:tamua_compact}
\end{figure*}
\FloatBarrier

\subsection{Long-Term Conversational Memory Prompts}

Figures~\ref{fig:realmem_prompt_a} and~\ref{fig:realmem_prompt_b} present the RealMem HMRU-I and RealMem HMRU-T prompt templates, respectively. HMRU-I presents memories by relevance, whereas HMRU-T organizes retrieved turns chronologically to support temporal state tracking.

\begin{figure*}[p]
\centering
\begin{realmemPrompt}{
    RealMem HMRU-I: Task-Aware Memory Utilization
}

\textbf{System Instruction}

You are the user's long-term personal assistant
(fitness, health, mental wellbeing, learning,
travel and other personal plans).
You talk with the same user over many sessions
spread across weeks. You cannot see the earlier
sessions directly; instead you receive memories
retrieved from them.

\medskip
\textbf{Retrieved Conversation Memory}

The following memories are sorted from most to
least relevant to the current message.

\begin{tcolorbox}[
    colback=gray!8,
    colframe=gray!35,
    boxrule=0.4pt,
    arc=4pt
]
\textbf{memory content:}
\rvar{stored memory}

\textbf{memory context:}
\rvar{generated context}

\textbf{memory keywords:}
\rvar{keywords}

\textbf{memory tags:}
\rvar{tags}

\textbf{talk start time:}
\rvar{timestamp}

\medskip
\textit{Additional retrieved memories follow.}
\end{tcolorbox}

\textbf{Current Session Date:}
\rvar{session date}

\textbf{Current User Message:}
\rvar{question}

\medskip
\textbf{Instructions}

\begin{enumerate}[
    leftmargin=*,
    itemsep=2pt,
    topsep=3pt
]

\item Each memory is one earlier conversation
message. ``Speaker user says'' identifies the
user, and ``Speaker assistant says'' identifies
the assistant. The session date is given by
``talk start time''. Context, keywords and tags
are automatically generated; use them for
interpretation but rely on the original content
for factual information.

\item Keep your thinking short. Identify which
memories matter and the latest state, then reply
directly to the user. Do not mention memories.

\item Ground the response in the user's own
goals, plans, numbers, schedules, names,
constraints and previously agreed decisions.
Avoid generic advice.

\item Continue the ongoing conversation.
If the assistant previously proposed a plan,
list or next step, build on it rather than
starting over.

\item When memories conflict, prefer the
information from the latest session date.
Treat earlier conflicting values as outdated.

\item Respect previously stated preferences,
dislikes, limits and conditions, even when
the current message does not repeat them.

\item For temporal questions, use the memory
session dates and current session date to
determine event order and elapsed time.

\item Do not invent personal information.
If required information is missing, briefly
acknowledge the limitation and provide
the best available assistance.

\item Answer every part of the current
message completely but concisely, without
unrelated content.

\end{enumerate}

\textbf{Your Reply}

\rvar{assistant response}

\end{realmemPrompt}

\caption{
\textbf{RealMem HMRU-I: relevance-ordered memory utilization.}
Retrieved memories retain their original A-MEM
representation, with task-aware instructions
for personalized, temporally consistent responses.
}
\label{fig:realmem_prompt_a}
\end{figure*}
\FloatBarrier

\begin{figure*}[p]
\centering
\begin{realmemPrompt}{
    RealMem HMRU-T: Temporal Memory and State Tracking
}

\textbf{System Instruction}

You are the user's long-term personal assistant
(fitness, health, mental wellbeing, learning,
travel and other personal plans).
You talk with the same user over many sessions
spread across weeks. You cannot see the earlier
sessions directly; instead you receive memories
retrieved from them.

\medskip
\textbf{Earlier Conversation}

Retrieved conversation turns are grouped by
session date from oldest to newest.
Within each session, the original turn order
is preserved.

\begin{tcolorbox}[
    colback=gray!8,
    colframe=gray!35,
    boxrule=0.4pt,
    arc=4pt
]
\textbf{Session:} \rvar{earlier session date}

\textbf{[R\#]} User:
\rvar{earlier user message}

\textbf{Summary:}
\rvar{memory context}

\medskip
\texttt{...}

\medskip
\textbf{[R\#]} Assistant:
\rvar{earlier assistant response}

\textbf{Summary:}
\rvar{memory context}

\medskip
\textbf{Session:} \rvar{latest session date}
\textbf{(Most Recent Session)}

\textbf{[R\#]} User:
\rvar{recent user message}

\textbf{Summary:}
\rvar{memory context}

\medskip
\textit{Additional retrieved turns may appear.}
\end{tcolorbox}

\textbf{Current Session Date:}
\rvar{session date}

\textbf{Current User Message:}
\rvar{question}

\medskip
\textbf{Instructions}

\begin{enumerate}[
    leftmargin=*,
    itemsep=2pt,
    topsep=3pt
]

\item The retrieved conversation is a partial
transcript organized chronologically.
``User'' and ``Assistant'' identify speakers;
``...'' marks omitted turns.
[R\#] indicates retrieval relevance,
with R1 being the most relevant memory.
Generated summaries provide context;
rely on message content for facts.

\item Before answering, briefly identify
three aspects of the user's conversational
state:

\begin{tcolorbox}[
    colback=gray!8,
    colframe=gray!35,
    boxrule=0.4pt,
    arc=4pt
]
\textbf{THREAD:}
\rvar{most recent discussion and next step}

\textbf{PROFILE:}
\rvar{relevant stable facts, goals,
preferences and constraints}

\textbf{UPDATES:}
\rvar{changed information and latest values}
\end{tcolorbox}

Keep these three statements short and
then proceed to the response.

\item Reply directly to the user as the
next message in the ongoing conversation.
Do not mention memories, the transcript
or the state checklist.

\item Continue the THREAD rather than
restarting the conversation. Respect
the PROFILE and use the latest values
identified in UPDATES.

\item Ground the response in the user's
own plans, numbers, schedules, names,
constraints and prior agreements,
rather than providing generic advice.

\item For temporal questions, use
historical session dates and the
current session date.

\item Do not invent personal facts.
If necessary information is missing,
acknowledge it briefly and provide
the best available assistance.

\item Answer every part of the current
message clearly and concisely.

\end{enumerate}

\textbf{Your Reply}

\rvar{assistant response}

\end{realmemPrompt}

\caption{
\textbf{RealMem HMRU-T: chronological memory organization and state tracking.}
The same retrieved memories are reorganized
by session date and original turn order,
with a compact state checklist for dialogue
continuity, user preferences and temporal updates.
}
\label{fig:realmem_prompt_b}
\end{figure*}
\FloatBarrier

\section{Additional Related Work}
\label{sec:additional-related-work}

\paragraph{Retrieval-augmented knowledge integration.}
Retrieval-augmented generation (RAG) conditions language-model predictions on evidence selected from external collections, providing an alternative to relying exclusively on parametric knowledge~\citep{lewis2020retrieval,karpukhin2020dense}. Subsequent work has incorporated retrieval and evidence assessment into the generation process itself~\citep{asai2024self}. In biomedical applications, retrieved demonstrations have supported few-shot medical entity recognition~\citep{li2024rt}, while external evidence has been used for healthcare prediction and cancer-related language understanding~\citep{li2025biomedrag,li2026cancerllm}. Evaluations of biomedical RAG have also examined robustness to noisy evidence and the ability of models to recognize their own limitations~\citep{li2025benchmarking}. Complementing external retrieval, K2K investigates access to knowledge within model representations through activation-guided probing and reranking~\citep{li2026efficient}. These studies establish a range of mechanisms for accessing task-relevant information, but evidence availability alone does not determine whether the information is appropriate for a particular clinical question.

\paragraph{Structured and relational evidence retrieval.}
Biomedical evidence frequently involves relationships between entities, events, and attributes that cannot be adequately represented by isolated text passages. GraphRAG~\citep{edge2024local} organizes entity relationships into hierarchical community summaries for corpus-level reasoning, while knowledge-graph retrieval has also been explored for tracking changing entity states~\citep{li2023understand}. HippoRAG~\citep{gutierrez2024hipporag} leverages knowledge graphs and graph-based ranking for multi-hop retrieval, whereas condensed relational paths have been used to support zero-shot link prediction~\citep{li2024condensed}. G-Retriever~\citep{he2024g} identifies relevant subgraphs for question answering over textual graphs, and KG2RAG~\citep{zhu2025knowledge} connects related passages through knowledge-graph structures to improve evidence coverage. LightRAG-Driven Medical QA~\citep{kushwaha2025lightrag} further explores graph-based retrieval for medical question answering. RiTeK~\citep{huang2026ritek} evaluates complex biomedical reasoning over textual knowledge graphs, while GLEN-Bench~\citep{huang2026glen} extends graph--language evaluation to nutritional-health knowledge. These studies demonstrate the value of relational structures for evidence retrieval, while leaving open how reliably retrieved information is preserved, updated, and utilized within persistent memory systems.

\paragraph{Adaptive retrieval and interaction-aware reasoning.}
Recent approaches increasingly integrate retrieval into multi-step reasoning rather than treating it as a fixed preprocessing operation. IRCoT~\citep{trivedi2023interleaving} interleaves retrieval with chain-of-thought reasoning, while FLARE~\citep{jiang2023active} introduces active retrieval during generation. RAG-GD~\citep{li2026context} explores in-context optimization to adapt how frozen generators utilize retrieved evidence, whereas DeepRAG~\citep{guan2026deeprag} investigates when additional retrieval is warranted during reasoning. RIVO~\citep{li2026rivo} examines lightweight interventions that redirect agent behavior during longer research trajectories, while Adaptive-RAG~\citep{jeong2024adaptive} selects retrieval strategies according to question complexity. RICE-PO~\citep{li2026rice} develops localized learning signals for interactive retrieval policies by evaluating executable retrieval actions and their counterfactual outcomes, whereas Self-RAG~\citep{asai2024self} incorporates self-reflection to guide retrieval and generation. Together, these methods address complementary aspects of adaptive reasoning, including evidence acquisition, generator adaptation, trajectory control, and policy learning, motivating the evaluation of retrieval as part of a broader information-use process.

\paragraph{Implications for biomedical agentic memory.}
Unlike one-off retrieval, persistent memory systems may transform source records into compact representations, revise those representations across interactions, and reuse them for subsequent questions. This setting raises additional questions about whether stored content remains supported by its source, whether retrieval identifies evidence relevant to the specific biomedical fact or clinical outcome, and whether the generator uses that evidence correctly. BioMem examines these dimensions jointly across biomedical reasoning, prediction, and long-term conversational tasks. HMRU further investigates whether complementary retrieval and task-aware evidence utilization can mitigate selected limitations without updating the underlying language-model parameters.

\end{document}